\documentclass[10pt,twocolumn,letterpaper]{article}

\usepackage{cvpr}
\usepackage{times}
\usepackage{epsfig}
\usepackage{graphicx}
\usepackage{amsmath}
\usepackage{amssymb}

\makeatletter
\newcommand*{\rom}[1]{\expandafter\@slowromancap\romannumeral #1@}
\makeatother

\usepackage[pagebackref=true,breaklinks=true,letterpaper=true,colorlinks,bookmarks=false]{hyperref}

\cvprfinalcopy 

\ifcvprfinal\fi
\begin{document}

\title{Towards Better Generalization: Joint Depth-Pose Learning without PoseNet}

\author{
Wang Zhao \quad Shaohui Liu \quad Yezhi Shu \quad Yong-Jin Liu\thanks{Corresponding author.} \\
Department of Computer Science and Technology, Tsinghua University, Beijing, China \\
\tt \small zhao-w19@mails.tsinghua.edu.cn,
b1ueber2y@gmail.com, \\
\tt \small shuyz19@mails.tsinghua.edu.cn,
liuyongjin@tsinghua.edu.cn
}

\maketitle
\thispagestyle{empty}

\begin{abstract}
   
In this work, we tackle the essential problem of scale inconsistency for self-supervised joint depth-pose learning. Most existing methods assume that a consistent scale of depth and pose can be learned across all input samples, which makes the learning problem harder, resulting in degraded performance and limited generalization in indoor environments and long-sequence visual odometry application. To address this issue, we propose a novel system that explicitly disentangles scale from the network estimation. Instead of relying on PoseNet architecture, our method recovers relative pose by directly solving fundamental matrix from dense optical flow correspondence and makes use of a two-view triangulation module to recover an up-to-scale 3D structure. Then, we align the scale of the depth prediction with the triangulated point cloud and use the transformed depth map for depth error computation and dense reprojection check. Our whole system can be jointly trained end-to-end. Extensive experiments show that our system not only reaches state-of-the-art performance on KITTI depth and flow estimation, but also significantly improves the generalization ability of existing self-supervised depth-pose learning methods under a variety of challenging scenarios, and achieves state-of-the-art results among self-supervised learning-based methods on KITTI Odometry and NYUv2 dataset. Furthermore, we present some interesting findings on the limitation of PoseNet-based relative pose estimation methods in terms of generalization ability. Code is available at \href{https://github.com/B1ueber2y/TrianFlow}{\color{cyan}{https://github.com/B1ueber2y/TrianFlow}}.

\vspace{-5pt}

\end{abstract}

\section{Introduction}

\begin{figure}[tb]
\begin{tabular}{cc}
{\includegraphics[width=0.55\linewidth, height=140pt]{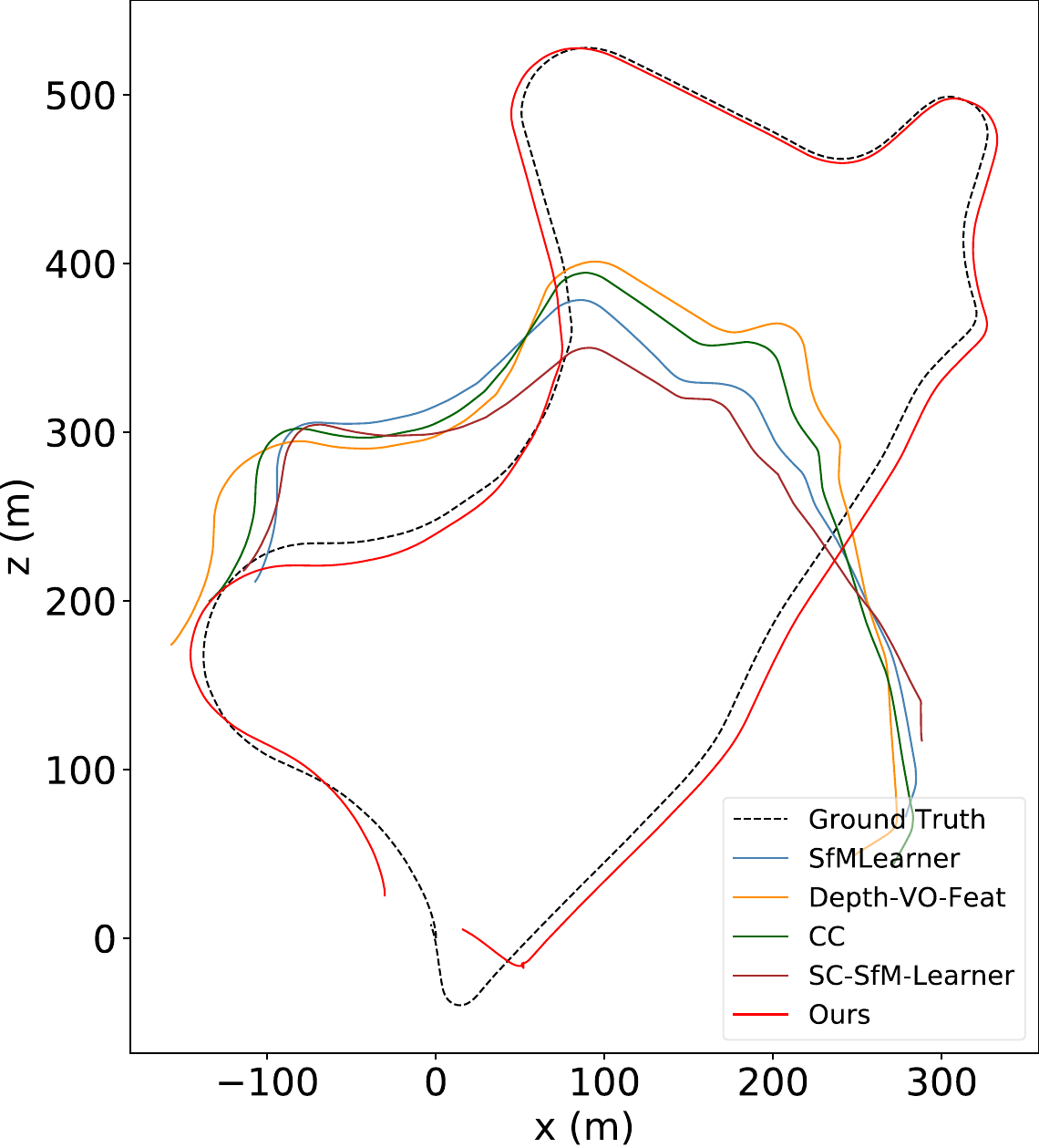}} &
{\includegraphics[width=0.35\linewidth, height=140pt]{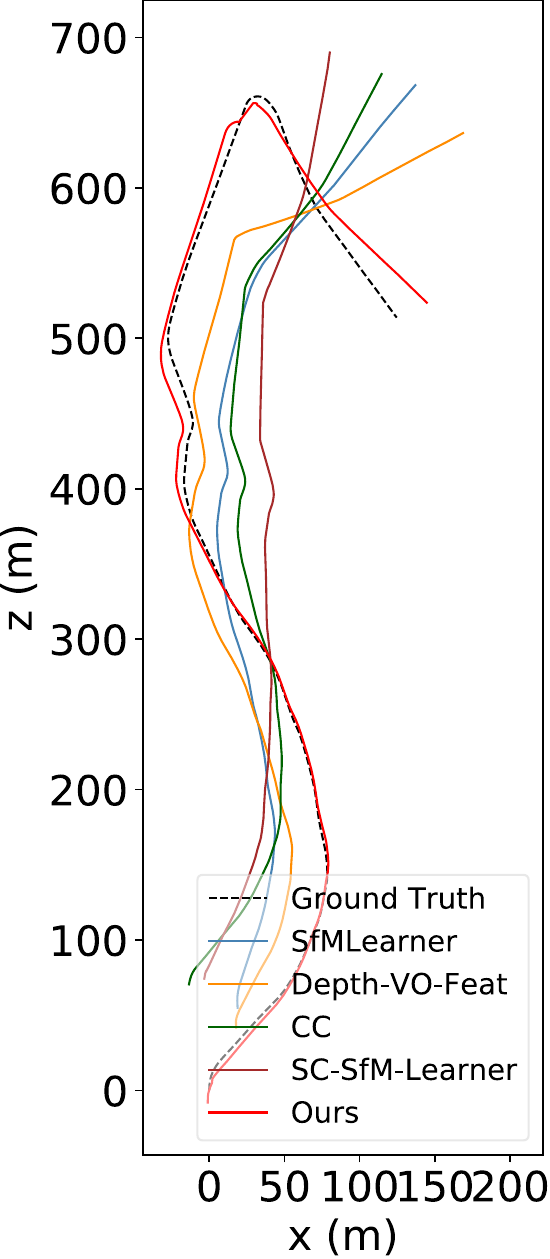}} \\
\end{tabular}
\centering
\caption{Visual odometry results on sampled sequence 09 and 10 from KITTI Odometry dataset. We sample the original sequences with large stride (stride=3) to simulate fast camera ego-motion that is unseen during training. Surprisingly, all tested PoseNet-based methods get similar failure on trajectory estimation under this challenging scenario. Our system significantly improves the generalization ability and robustness and still works reasonably well on both sequences. See more discussions in Sec \ref{sec::exp_unseen}.}
\label{fig::firstpage}
\vspace{-15pt}
\end{figure}

Reconstructing the underlying 3D scenes from a collection of video frames or multi-view images has been a long-standing fundamental topic named structure-from-motion (SfM), which serves as an essential module to many real-world applications such as autonomous vehicles, robotics, augmented reality, etc. While traditional methods are built on the golden rule of feature correspondence and multi-view geometry, a recent trend of deep learning based methods \cite{roy2016monocular,godard2017unsupervised,zhou2017unsupervised} try to jointly learn the prediction of monocular depth and ego-motion in a self-supervised manner, aiming to make use of the great learning ability of deep networks to learn geometric priors from large amount of training data. 

The key to those self-supervised learning methods is to build a task consistency for training separated CNN networks, where depth and pose predictions are jointly constrained by depth reprojection and image reconstruction error.
While achieving fairly good results, most existing methods assume that a consistent scale of CNN-based monocular depth prediction and relative pose estimation can be learned across all input samples, since relative pose estimation inherently has scale ambiguity. Although several recent proposals manage to mitigate this scale problem \cite{bian2019unsupervised, garg2019learning}, this strong hypothesis still makes the learning problem difficult and leads to severely degraded performance, especially in long-sequence visual odometry applications and indoor environments, where the changes of relative pose across sequences are significantly remarkable. 

Motivated by those observations, we propose a new self-supervised depth-pose learning system which explicitly disentangles scale from the joint estimation of the depth and relative pose. Instead of using a CNN-based camera pose prediction module (e.g. PoseNet), we directly solve the fundamental matrix from optical flow correspondences and implement a differentiable two-view triangulation module to locally recover an up-to-scale 3D structure. 
This triangulated point cloud is later used to align the predicted depth map via a scale transformation for depth error computation and reprojection consistency check.

Our system essentially resolves the scale inconsistency problem in design. With two-view triangulation and explicit scale-aware depth adaptation, the scale of the predicted depth always matches that of the estimated pose, enabling us to remove the scale ambiguity for joint depth-pose learning. Likewise, we borrow the advantage of traditional two-view geometry to acquire more direct, accurate and robust depth supervision in a self-supervised end-to-end manner, where the depth and flow prediction can benefit from each other. Moreover, because our relative pose is directly solved from the optical flow, we simplify the learning process and do not require the knowledge of correspondence to be learned from the PoseNet architecture, enabling our system to have better generalization ability in challenging scenarios. See an example in Figure \ref{fig::firstpage}.

Experiments show that our unified system significantly improves the robustness of self-supervised learning methods in challenging scenarios such as long video sequences, unseen camera ego-motions, and indoor environments. Specifically, our proposed method achieves significant performance gain on NYU v2 dataset and KITTI Odometry over existing self-supervised learning-based methods, and maintains state-of-the-art performance on KITTI depth and flow estimation. We further test our framework on TUM-RGBD dataset and again demonstrate its much promising generalization ability compared to baselines.

\section{Related Work}


\paragraph{Monocular Depth Estimation.}
Recovering 3D depth from a single monocular image is a fundamental problem in computer vision. Early methods \cite{saxena2006learning,saxena2008make3d} use feature vectors along with a probabilistic model to provide monocular clues. Later, with the advent of deep networks, a variety of systems \cite{eigen2014depth,fu2018deep,roy2016monocular} are proposed to learn monocular depth estimation from groundtruth depth maps in a supervised manner. To resolve the data deficiency problem, \cite{mayer2018makes} uses synthetic data to help the disparity training, and several works \cite{li2018megadepth,klodt2018supervising,li2019learning,lasinger2019towards} leverage standard structure-from-motion (SfM) pipeline \cite{schoenberger2016SfM,schoenberger2016mvs} to generate a psuedo-groundtruth depth map by reprojecting the reconstructed 3D structure. Recently, a bunch of works \cite{garg2016unsupervised,godard2017unsupervised,zhou2017unsupervised} on self-supervised learning are proposed to jointly estimate other geometric entities that help depth estimation learning via photometric reprojection error. However, although some recent works \cite{wang2018learning,garg2019learning} try to address the scale ambiguity for monocular depth estimation with either normalization or affine adaptation, self-supervised methods still suffer from the problem of scale inconsistency when applied to challenging scenarios. Our work combines the advantages of SfM-based unsupervised methods and self-supervised learning methods, essentially disentangles scale from our learning process and benefits from the more accurate and robust triangulated structure with two-view geometry.

\vspace{-5pt}
\paragraph{Self-Supervised Depth-Pose Learning.}
Struction-from-motion (SfM) is a golden standard for depth reconstruction and camera trajectory recovery from videos and image collections. Recently many works \cite{tateno2017cnn,bloesch2018codeslam,yang2018deep,tang2018ba} try to combine neural networks into SfM pipeline to make use of the learned geometric priors from training data. Building on several unsupervised methods \cite{garg2016unsupervised,godard2017unsupervised}, Zhou \etal \cite{zhou2017unsupervised} first proposes a joint unsupervised learning framework of depth and camera ego-motion from monocular videos. The core idea is to use photometric error as supervision signal to jointly train depth and ego-motion networks. Along this line, several methods \cite{yin2018geonet, zou2018df, mahjourian2018unsupervised, bian2019unsupervised, ranjan2019competitive, luo2018every, casser2019depth, chen2019self} further improve the performance by incorporating better training strategies and additional constraints including ICP regularization \cite{mahjourian2018unsupervised}, collaborative competition \cite{ranjan2019competitive}, dense online bundle adjustment \cite{casser2019depth,chen2019self}, etc. Most related to us, Bian \etal \cite{bian2019unsupervised} introduce geometry consistency loss to enforce the scale-consistent depth learning. Different from them, our method essentially avoids the scale inconsistency in deign by directly solving relative pose from optical flow correspondence. Our system designs and findings are orthogonal to existing depth-pose learning works, significantly improving those methods on both accuracy and generalization.

\begin{figure*}[tb]
{\includegraphics[width=0.95\linewidth]{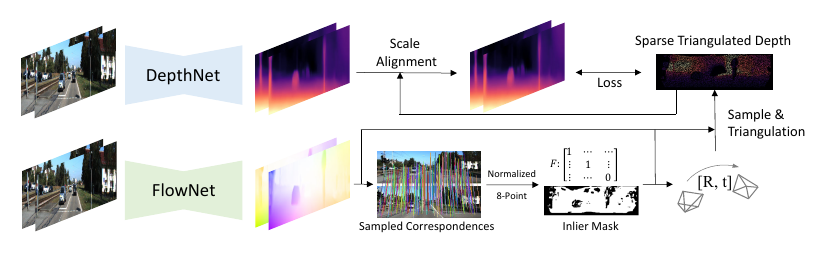}}
\centering
\caption{System overview. DepthNet takes each input image and predicts monocular depths respectively. FlowNet take image pairs as input and predict optical flows. The relative pose is recovered by sampling correspondences, solving the fundamental matrix, and cheirality condition check. Accurate pixel matches are re-sampled and used for triangulation. Depth predictions are aligned according to sparse triangulation depth, and then losses are measured respectively, to supervise DepthNet and FlowNet jointly.}
\label{fig::pipeline}
\vspace{-10pt}
\end{figure*}

\vspace{-5pt}
\paragraph{Two-view Geometry.}

Establishing pixel-wise correspondences between two images is a long-standing visual problem. Traditional methods utilize hand-crafted descriptors \cite{lowe1999object,bay2006surf,rublee2011orb} to build rough correspondence for the subsequent fundamental matrix estimation. Recently, building on classic works of optical flow \cite{horn1981determining,lucas1981iterative}, researchers \cite{dosovitskiy2015flownet, ilg2017flownet, sun2018pwc} find deep neural networks powerful on feature extraction and dense correspondence estimation between adjacent frames. Likewise, several self-supervised methods \cite{janai2018unsupervised, meister2018unflow,liu2019selflow} are proposed to supervise optical flow training with photometric consistency. 

Another line of research is to combine learning-based methods with the fundamental matrix estimation after establishing the correspondence. While some researches \cite{brachmann2017dsac,ranftl2018deep} focus on making RANSAC \cite{fischler1981random} differentiable, another alternative is to use an end-to-end pose estimation network \cite{kendall2015posenet}. However, some recent findings \cite{sattler2019understanding,zhou2019learn} on image-based localization show that PoseNet design \cite{kendall2015posenet} can degrade the generalization ability compared to geometry-based methods. Also, the inherent problem of scale ambiguity for pose estimation makes it hard to decouple with depth scale during joint training. In our work, we show that by building on conventional two-view geometry, our optical flow estimation module is able to accurately recover relative poses and can benefit from the joint depth-pose learning.

\section{Method}

\subsection{Motivation and System Overview}

The central idea of existing self-supervised depth-pose learning methods is to learn two separated networks on the estimation of monocular depth and relative pose by enforcing geometric constraints on image pairs. Specifically, the predicted depth is reprojected onto another image plane using the predicted relative camera pose and then photometric error is measured. However, this class of methods assume a consistent scale of depth and pose across all images, which could make the learning problem difficult and lead to a scale drift when applied to visual odometry applications. 

Some recent proposals \cite{wang2018learning,bian2019unsupervised} introduce additional consistency constraints to mitigate this scale problem. Nonetheless, the scale-inconsistent issue naturally exists because the scales of the estimated depth and pose from neural networks are hard to measure. Also, the photometric error on the image plane supervises the depth in an implicit manner, which could suffer from data noise when large textureless regions exist. Furthermore, similar to two recent findings \cite{sattler2019understanding,zhou2019learn} that CNN-based absolute pose estimation is difficult to generalize beyond image retrieval, the performance of the CNN-based ego-motion estimation also significantly degrades when applied to challenging scenarios.

To address the above challenges, we propose a novel system that explicitly disentangles scale consistency at both training and inference. The overall pipeline of our method is shown in Figure \ref{fig::pipeline}. Instead of relying on CNN-based relative pose estimation, we first predict optical flow and solve the fundamental matrix from the dense flow correspondence, thereby recovering relative camera pose. Then, we sample over the inlier regions and use a differentiable triangulation module to reconstruct an up-to-scale 3D structure. Finally, depth error is directly computed after a scale adaptation from the predicted depth to the triangulated structure and reprojection error on depth and flow is measured to further enforce end-to-end joint training. Our training objective $L$ is formulated as follows:

\begin{equation}
L=w_{1}L_{f} + w_2L_{d} + w_3L_{p} + w_4L_{s}.
\end{equation}

The $L_{f}$ denotes the unsupervised loss on optical flow, where we follow the photometric error design (pixel + SSIM \cite{wang2004image} + smooth) on PWC-Net \cite{sun2018pwc}. Occlusion mask $M_o$ is derived from optical flow by following \cite{wang2018occlusion}. We also add a forward-backforward consistency \cite{yin2018geonet} to generate a score map $M_s$ for subsequent fundamental matrix estimation. $L_{d}$ is the loss between triangulated depth and predicted depth. $L_{p}$ is the reprojection error for image pairs, which consists of two parts, depth map reconstruction error and flow error between optical flow and rigid flow generated by depth reprojection. $L_{s}$ is the depth smoothness loss, which follows the same formulation in \cite{bian2019unsupervised}.

In the following parts, we first describe how we recover relative pose via fundamental matrix from optical flow. Then, we show how to use the recovered pose to build up self supervision geometrically without scale ambiguity. Finally, a brief description is given on the inference pipeline of our system when applied to visual odometry applications.

\subsection{Fundamental Matrix from Correspondence}
We recover camera pose from optical flow correspondence via traditional fundamental matrix computation algorithm. Optical flow offers correspondence for every pixel, while some of them are noisy and thus not suitable for solving the fundamental matrix. We first select reliable correspondences using the occlusion mask $M_o$ and forward-backward flow consistency score map $M_s$, which are both generated from our flow network. Specifically, we sample the correspondences that locate in non-occluded regions and have top 20\% forward-backward scores. Then we randomly acquire 6k samples out of the selected correspondences and solve the fundamental matrix $F$ via the simple normalized 8-point algorithm \cite{hartley1997defense} in RANSAC \cite{fischler1981random} loop. Fundamental matrix is then decomposed into camera relative pose, which is denoted as $[R, t]$. Note that there are 4 possible solutions for $[R, t]$ and we adopt cheirality condition check, meaning that the triangulated 3D points must be in front of both cameras, to find the best one solution. In this way, our predicted camera pose fully depends on the optical flow network, which can better generalize across image sequences and under challenging scenarios.

\subsection{Two-view Triangulation as Depth Supervision}
\label{sec::3.3}
Recovering the relative camera pose with fundamental matrix estimation from optical flow formulates an easier learning problem and improves the generalization, but cannot enforce scale-consistent prediction on its own. To follow up with this design, we propose to explicitly align the scale of depth and pose. Intuitively two reasonable solutions on scale optimization exist: 1) aligning depth with pose 2) aligning pose with depth. We adopt the former one as it can be formulated as a linear problem using two-view triangulation \cite{hartley1997triangulation}.

\begin{figure}[tb]
\setlength\tabcolsep{2pt} 
\begin{tabular}{cc}
\includegraphics[width=0.45\linewidth]{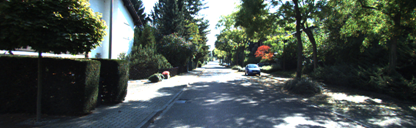} & 
\includegraphics[width=0.45\linewidth]{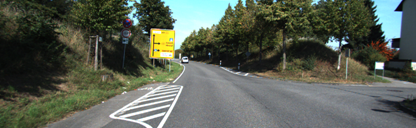} \\
\includegraphics[width=0.45\linewidth]{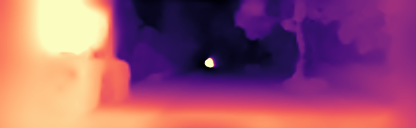} &
\includegraphics[width=0.45\linewidth]{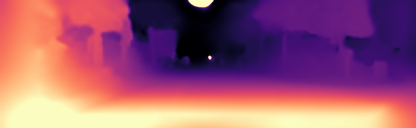} \\

\end{tabular}
\centering
\caption{Dense triangulation examples. While most of the triangulated matches are pretty good, the depth values around occluded areas and epipole regions are noisy. In these two examples, epipoles locate near the image center and the nearby triangulated depths are negative or very close to zero. Thus we only use sampled sparse accurate triangulation depth as supervision.}
\label{fig::triangulation}
\vspace{-10pt}
\end{figure}

Again, instead of using all pixel matches to perform dense triangulation, we first select top accurate correspondences. Specifically, we generate an inlier score map $M_r$ by computing the distance map $D_{epi}$ from each pixel to its corresponding epipolar line, which is helpful for masking out bad matches and non-rigid regions, such as moving objects. Then this inlier score map $M_r$ is combined with occlusion mask $M_o$, optical flow forward-backward score $M_s$, to sample rigid, non-occluded and accurate correspondences. Here we also randomly acquire 6k samples out of the top 20\% correspondence and perform two-view triangulation to reconstruct an up-to-scale 3D structure. 
We adopt the mid-point triangulation as it has a linear and robust solution. Its formulation is as follows:

\begin{equation}
x^* = \underset{x}{\operatorname{argmin}}{\ [d(L_1, x)]^2 + [d(L_2, x)]^2},
\label{eq::mid-point}
\end{equation}

\noindent where $L_1$ and $L_2$ denote two camera rays generated from optical flow correspondence. This problem can be directly solved analytically and the solver is naturally differentiable, enabling our system to perform end-to-end joint training. The derivation of its analytical solution is included in supplementary materials. We use the triangulated 3D structure as the depth supervision. To mitigate the numerical issue, such as triangulation of matches around epipoles, we filter the correspondence online with respect to the angle of the camera rays. Also, we filter the triangulated samples with negative or out-of-bound depth reprojection. 
Figure \ref{fig::triangulation} visualizes samples for the depth reprojection of the dense triangulated structure. The quality of the depth is much promising and feasible to be used as a psuedo depth groundtruth signal to guide the network learning. This design shares similar spirits with many recent methods \cite{li2018megadepth,klodt2018supervising,li2019learning,lasinger2019towards} on supervising the monocular depth estimation with offline SfM inference where they also use the reconstructed structure as the psuedo groundtruth. Compared to those works, our online robust triangulation module explicitly handles occlusion, moving objects and bad matches, and is successfully integrated into the joint training system where correspondence generation and depth prediction could benefit together.

\subsection{Scale-invariant Design}
As aforementioned, we can resolve the scale-inconsistent problem by aligning predicted depth with the triangulated structure. Specifically, we align the monocular depth estimation $D$ with a single scale transformation $s$ to minimize the error between the transformed depth $D_{t}=sD$ and the psuedo groundtruth depth $D_{tri}$ from triangulation in Eq. (\ref{eq::depth-error}). Then, the minimized error is used as the depth loss for back-propagation. This online fitting technique was also introduced in a recent work \cite{garg2019learning}.

\begin{equation}
L_{d} = (\frac{D_{tri} - D_t}{D_{tri}})^2
\label{eq::depth-error}
\end{equation}

The transformed depth is explicitly aligned to the triangulated 3D structure, whose scale is decided by relative pose scale, thus scale inconsistency is essentially disentangled from the system. Also, the transformed depth can be further used for computing the dense reprojection error $L_{p}$. This error is formulated in Eq. (\ref{eq::reproj}):

\begin{equation}
L_{p}=w_{31}L_{pf} + w_{32}L_{pd},
\label{eq::reproj}
\end{equation}

Given an image pair $(I_a, I_b)$, scale-transformed depth estimations $(D_a, D_b)$, camera intrinsic parameter $\rm K$, and recovered relative pose $\rm T_{ab}$ from optical flow $F_{ab}$, loss $L_{pf}$ is calculated as follows, which measures the 2D error between optical flow and rigid flow generated by depth reprojection.
\begin{equation}
\begin{aligned}
    p_{bd} &= \phi({\rm K}[{\rm T}_{ab}{D}_a(p_a){\rm K}^{-1}(h(p_a)])) \\
    p_{bf} &= p_a + F_{ab}(p_a) \\
    L_{pf} &= \frac{1}{\vert M_r\rvert}\sum_{p_a}{M_r(p_a)\lvert p_{bd} - p_{bf}\rvert} + \vert D_{epi}\rvert\\
    \label{eq::depth_reproj1}
\end{aligned}
\vspace{-15pt}
\end{equation}
\noindent where $p_a$ is the pixel coordinate $(x,y)$ in $I_a$, and $h(p_a)$ indicates the homogeneous coordinates of $p_a$. Operator $\phi([x,y,z])=[x/z, y/z]$ gives pixel coordinates. As mentioned in Sec \ref{sec::3.3}, $D_{epi}$ is the distance map of each pixel to its corresponding epipolar line and $M_r$ is the inlier score map. $\vert D_e\rvert$ serves as a geometric regularization term to help improve the correspondences. $\vert M_r\rvert = \sum_{p_a}{M_r(p_a)}$ is for normalization. Depth reprojection error $L_{pd}$ is defined as:
\begin{equation}
    L_{pd} = \frac{1}{\lvert M_o M_r\rvert} \sum_{p_a} { M_o(p_a)M_r(p_a) \lvert1 - \frac{D_b^a(p_{bd})}{D_b^s(p_{bd})} \rvert}
    \label{eq::depth_reproj2}
\end{equation}
\noindent where $D_b^a$ is the reprojected depth map by $D_a$ and $\rm T_{ab}$. $D_b^s$ is the interpolated depth map of $D_b$ to align with reprojected pixel coordinates $p_{bd}$, which is defined in Eq. (\ref{eq::depth_reproj1}). $M_o$ is the occlusion mask from optical flow. 

\subsection{Inference Pipeline on Video Sequences}
At inference step, we use the same strategy for relative pose estimation via fundamental matrix estimation from optical flow correspondence. Then, the scale of the triangulated structure is aligned as the same with that of monocular depth estimation. When the optical flow magnitude is too small, we use perspective-n-point (PnP) method over the predicted depth directly. In this way, we essentially avoid the scale inconsistency between depth and pose during inference. A recent paper \cite{zhan2019visual} employs similar visual odometry inference strategies to utilize neural network predictions. However, their depth and flow network are pre-trained separately using PoseNet architecture, while our method builds a robust joint learning system to learn better depth, pose and flow predictions in a self-supervised manner.

\section{Experiments}

\subsection{Implementation Details}
\begin{figure*}[tb]
\scriptsize
\setlength\tabcolsep{1.0pt} 
\renewcommand{\arraystretch}{1.0}
\begin{tabular}{ccccc}
{\includegraphics[width=0.19\linewidth]{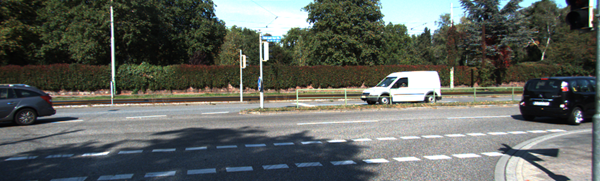}} &
{\includegraphics[width=0.19\linewidth]{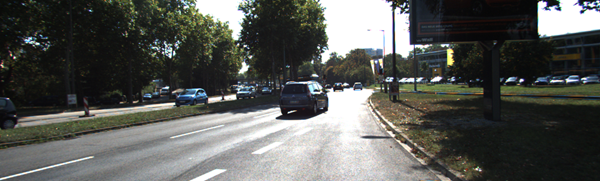}}
&
{\includegraphics[width=0.19\linewidth]{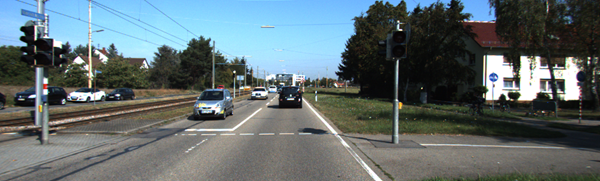}}
&
{\includegraphics[width=0.19\linewidth]{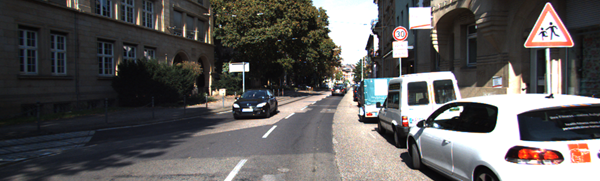}}
&
{\includegraphics[width=0.19\linewidth]{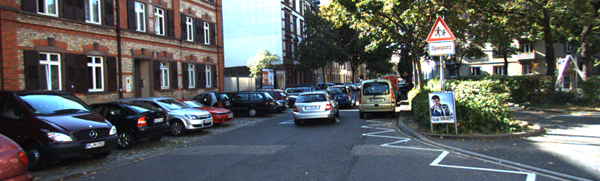}} \\

{\includegraphics[width=0.19\linewidth]{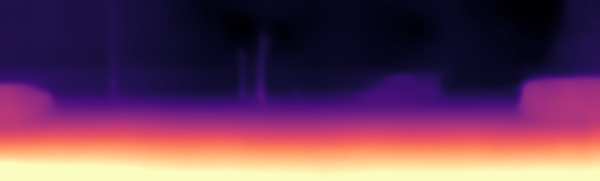}} &
{\includegraphics[width=0.19\linewidth]{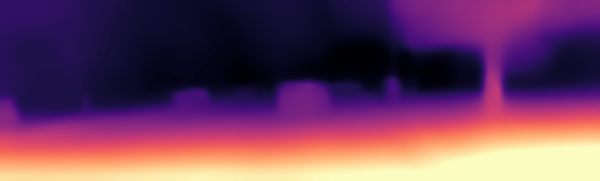}} &
{\includegraphics[width=0.19\linewidth]{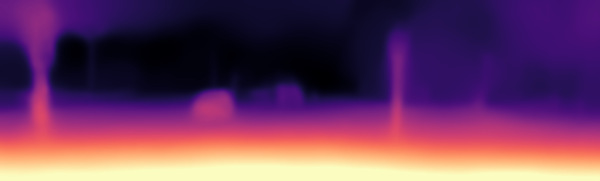}} &
{\includegraphics[width=0.19\linewidth]{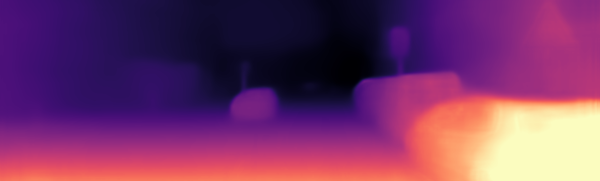}} &
{\includegraphics[width=0.19\linewidth]{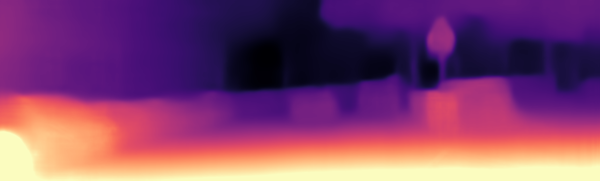}} \\

{\includegraphics[width=0.19\linewidth]{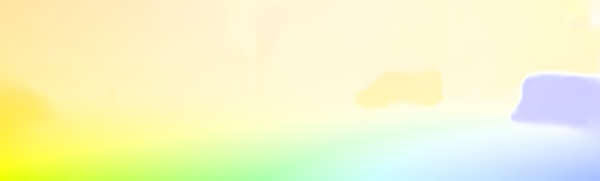}} &
{\includegraphics[width=0.19\linewidth]{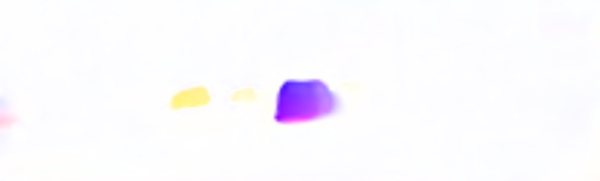}} &
{\includegraphics[width=0.19\linewidth]{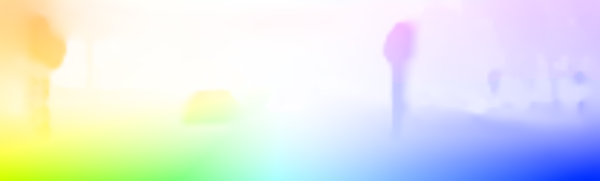}} &
{\includegraphics[width=0.19\linewidth]{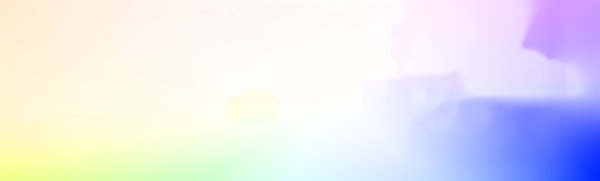}} &
{\includegraphics[width=0.19\linewidth]{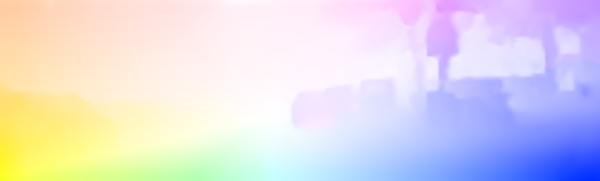}} \\

{\includegraphics[width=0.19\linewidth]{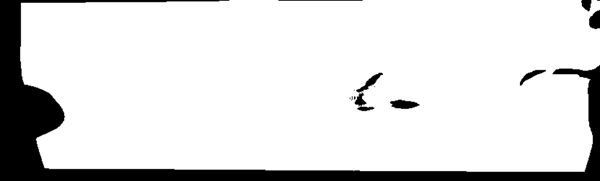}} &
{\includegraphics[width=0.19\linewidth]{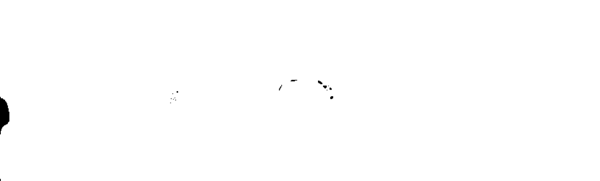}} &
{\includegraphics[width=0.19\linewidth]{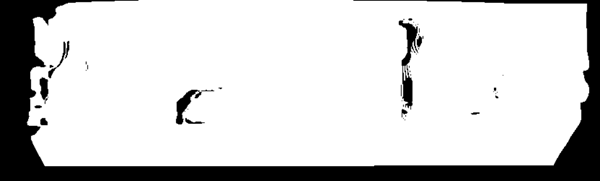}} &
{\includegraphics[width=0.19\linewidth]{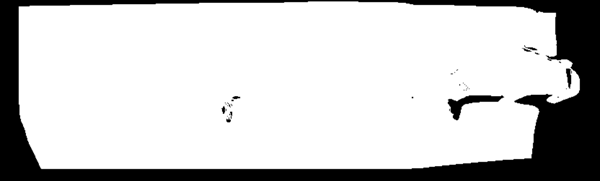}} &
{\includegraphics[width=0.19\linewidth]{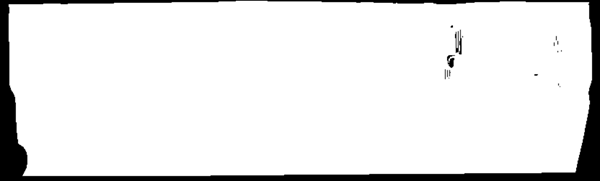}} \\
\end{tabular}
\centering
\vspace{0.05cm}
\caption{Qualitative results on KITTI dataset. \textbf{Top to bottom}: Original image, depth prediction, optical flow prediction and occlusion mask prediction.}
\label{fig::kitti_flow}
\vspace{-10pt}
\end{figure*}

\noindent
\textbf{Dataset.} We first validate our design on KITTI dataset \cite{geiger2012we}, then conduct extensive experiments on KITTI Odometry, NYUv2 \cite{silberman2012indoor} and TUM-RGBD \cite{sturm2012benchmark} datasets to demonstrate the robustness and generalization ability of our proposed system. For original KITTI dataset, we use Eigen \etal's split \cite{eigen2014depth} of the raw dataset for training, which is consistent with related works \cite{zhou2017unsupervised, ranjan2019competitive, chen2019self, godard2018digging}. The images are resized to 832$\times$256. We evaluate the depth network on the Eigen \etal's testing split, and the optical flow network on KITTI 2015 training set. For KITTI Odometry dataset, we follow the standard setting \cite{chen2019self, zhou2017unsupervised, yin2018geonet} of using sequences 00-08 for training and 09-10 for testing. Since the camera ego-motions in KITTI odometry dataset are relatively regular and steady, we sample the original test sequences to shorter versions, mimicking fast camera motions, for testing the generalization ability of networks on unseen data. NYUv2 \cite{silberman2012indoor} and TUM-RGBD \cite{sturm2012benchmark} are two challenging indoor datasets which consist of large textureless surfaces and more complex camera ego-motions.
~\\

\noindent
\textbf{Network Architectures.} Since our work focuses on an improved self-supervised depth-pose learning scheme, we adopt similar network designs that align with existing self-supervised learning methods. For the depth network, we use the same architecture as \cite{godard2018digging} which adopts ResNet18 \cite{he2016deep} as encoder and DispNet \cite{godard2017unsupervised} as decoder. The optical flow network is based on PWCNet \cite{sun2018pwc} and handles occlusion using the method described in \cite{wang2018occlusion}. Camera pose is calculated from filtered optical flow correspondences in a non-parametric manner.

~\\
\noindent
\textbf{Training.} Our system is implemented in PyTorch \cite{paszke2017automatic}. We use Adam \cite{kingma2014adam} optimizer and set learning rate to $10^{-4}$ and batch size to 8. The whole training schedule consists of three stages. Firstly, we only train optical flow network in an unsupervised manner via image reconstruction loss. After 20 epochs, we freeze optical flow network and train the depth network for another 20 epochs. Finally, we jointly train both networks for 10 epochs. 

\subsection{Conventional KITTI Setting}

\noindent\textbf{Monocular Depth Estimation.} We report results on monocular depth estimation on Eigen \etal's testing split on KITTI \cite{geiger2012we} dataset. The results are summarized in Table \ref{tab:depth}. Our method achieves comparable or better performance with state-of-the-art methods \cite{godard2018digging,gordon2019depth}. The performance gain is benefited from our system design, where the scale is disentangled from training and robust supervision is acquired from two-view triangulation module. We further explore the effects of different loss terms. The performance slightly drops without reprojection loss $L_p$ as shown in Table \ref{tab:depth}, and the training cannot converge without triangulation supervision loss $L_d$. Figure \ref{fig::kitti_flow} shows qualitative results of our depth prediction. 
Note that our method is orthogonal to many previous works, and could be potentially incorporated with many advanced techniques such as online refinement \cite{casser2019depth,chen2019self}, and more effective architecture \cite{guizilini2019packnet}.

\begin{table*}[t]
\begin{center}
\begin{tabular}{lcccc|ccc}
& \multicolumn{4}{c|}{Error} & \multicolumn{3}{c}{Accuracy, $\delta$ } \\ \cline{2-5}  \cline{6-8}
Method &AbsRel&SqRel&RMS&RMSlog &$<$$1.25$&$<$$1.25^2$& $<$$1.25^3$\\ 
\hline
{ Zhou \etal \cite{zhou2017unsupervised}} &0.183 & 1.595 & 6.709 & 0.270 & 0.734 & 0.902 & 0.959 \\
{ Mahjourian \etal \cite{mahjourian2018unsupervised}} & 0.163 & 1.240 & 6.220 & 0.250 & 0.762 & 0.916 & 0.968 \\
{ Geonet \cite{yin2018geonet} } &{0.155} & 1.296 & 5.857 & 0.233 & {0.793} & {0.931} & {0.973} \\
{ DDVO \cite{wang2018learning} } &{0.151} & 1.257 & 5.583 & 0.228 & 0.810 & {0.936} & 0.974 \\
{ DF-Net \cite{zou2018df} }  & 0.150  &1.124&  5.507& 0.223& 0.806& 0.933& 0.973 \\
{ CC \cite{ranjan2019competitive}} &  0.140 & 1.070 & 5.326 & 0.217 & 0.826 &   0.941 & 0.975 \\%
{ EPC++ \cite{luo2018every}} &  0.141 & 1.029 & 5.350 & 0.216 & 0.816 &   0.941 & 0.976 \\
{ Struct2depth (-ref.) \cite{casser2019depth}}  &  0.141 & 1.026 & 5.291 & 0.215 & 0.816 &   0.945 & 0.979 \\
{ GLNet (-ref.) \cite{chen2019self}} & 0.135 & 1.070 & 5.230 & 0.210 &
0.841 & 0.948 & 0.980 \\
{ SC-SfMLearner \cite{bian2019unsupervised}} &  0.137 & 1.089 & 5.439 & 0.217 & 0.830 & 0.942 & 0.975 \\
{ Gordon \etal \cite{gordon2019depth} } & 0.128 & 0.959 & 5.230 & 0.212 &
0.845 & 0.947 & 0.976 \\
{ Monodepth2 (w/o pretrain)~\cite{godard2018digging} } & 0.132 & 1.044 & 5.142 & 0.210 &
0.845 & 0.948 & 0.977 \\
{ Monodepth2$^\dag$ \cite{godard2018digging} }  & 0.115 & 0.882 & 4.701 & 0.190 & \textbf{0.879} & \textbf{0.961} & 0.982 \\

\hline
{ Ours (w/o pretrain and $L_p$)} & 0.135 & 0.932 & 5.128 & 0.208 & 0.830 & 0.943 & 0.978  \\
{ Ours (w/o pretrain)} & 0.130 & 0.893 & 5.062 & 0.205 &
0.832 & 0.949 & 0.981 \\
{ Ours$^\dag$} &  \textbf{0.113} & \textbf{0.704} & \textbf{4.581} & \textbf{0.184} & 0.871 & \textbf{0.961} & \textbf{0.984} \\
\hline
\end{tabular}
\vspace{0.1cm}
\caption{Quantitative comparison between our proposed system and state-of-the-art depth-pose learning methods (without post-processing) for monocular depth Estimation on KITTI \cite{geiger2012we} dataset. $^\dag$ indicates ImageNet pretraining.
}
\label{tab:depth}
\end{center}
\vspace{-20pt}
\end{table*}

~\\
\noindent\textbf{Optical Flow Estimation.} Table \ref{tab:flow} summarizes the results of optical flow estimation on KITTI 2015 training set. We also report the performance of only training our optical flow network, denoted as FlowNet-only. Results show that the optical flow module can benefit from joint depth-pose learning process and therefore outperforms most previous unsupervised flow estimation methods and joint learning methods. Figure \ref{fig::kitti_flow} shows some qualitative results.

\begin{table}[t]
\begin{center}
\begin{tabular}{lccc}
{Method} & {Noc} & {All} & {Fl}  \\ 
\hline
{FlowNetS \cite{ilg2017flownet} } & 8.12 & 14.19 & - \\
{FlowNet2 \cite{sun2018pwc} } & 4.93 & 10.06  & 30.37\% \\
\hline
{UnFlow \cite{meister2018unflow}} & - & 8.10 & 23.27\% \\
{Back2Future \cite{janai2018unsupervised} } & - & {7.04}  &  {24.21\%}\\\hline
{Geonet \cite{yin2018geonet}} & 8.05  & 10.81 & {-} \\
{DF-Net \cite{zou2018df} } & - & 8.98 & 26.01\% \\
{EPC++ \cite{luo2018every}} & - & 5.84 & - \\
{CC \cite{ranjan2019competitive}} & - & \textbf{5.66} & 20.93\% \\
{GLNet \cite{chen2019self}} & 4.86 & 8.35 & - \\
{Ours (FlowNet-only)} & 4.96 & 8.97 & 25.84\%\\
{Ours} & \textbf{3.60} & 5.72 & \textbf{18.05\%}\\
\hline
\end{tabular}
\vspace{0.08cm}
\caption{Optical flow estimation results. We report the average end-point-error (EPE) on non-occluded regions and overall regions, and Fl score on KITTI 2015 training set, following \cite{yin2018geonet, chen2019self}. Top 2 rows: supervised methods which are trained on synthetic data only. Middle 2 rows: unsupervised optical flow learning methods. Bottom rows: joint depth-pose learning methods.}
\label{tab:flow}
\end{center}
\vspace{-20pt}
\end{table}

\subsection{Generalization on Long Sequences}
\begin{figure}[tb]
\scriptsize
\begin{tabular}{cc}
{\includegraphics[width=0.55\linewidth,height=120pt]{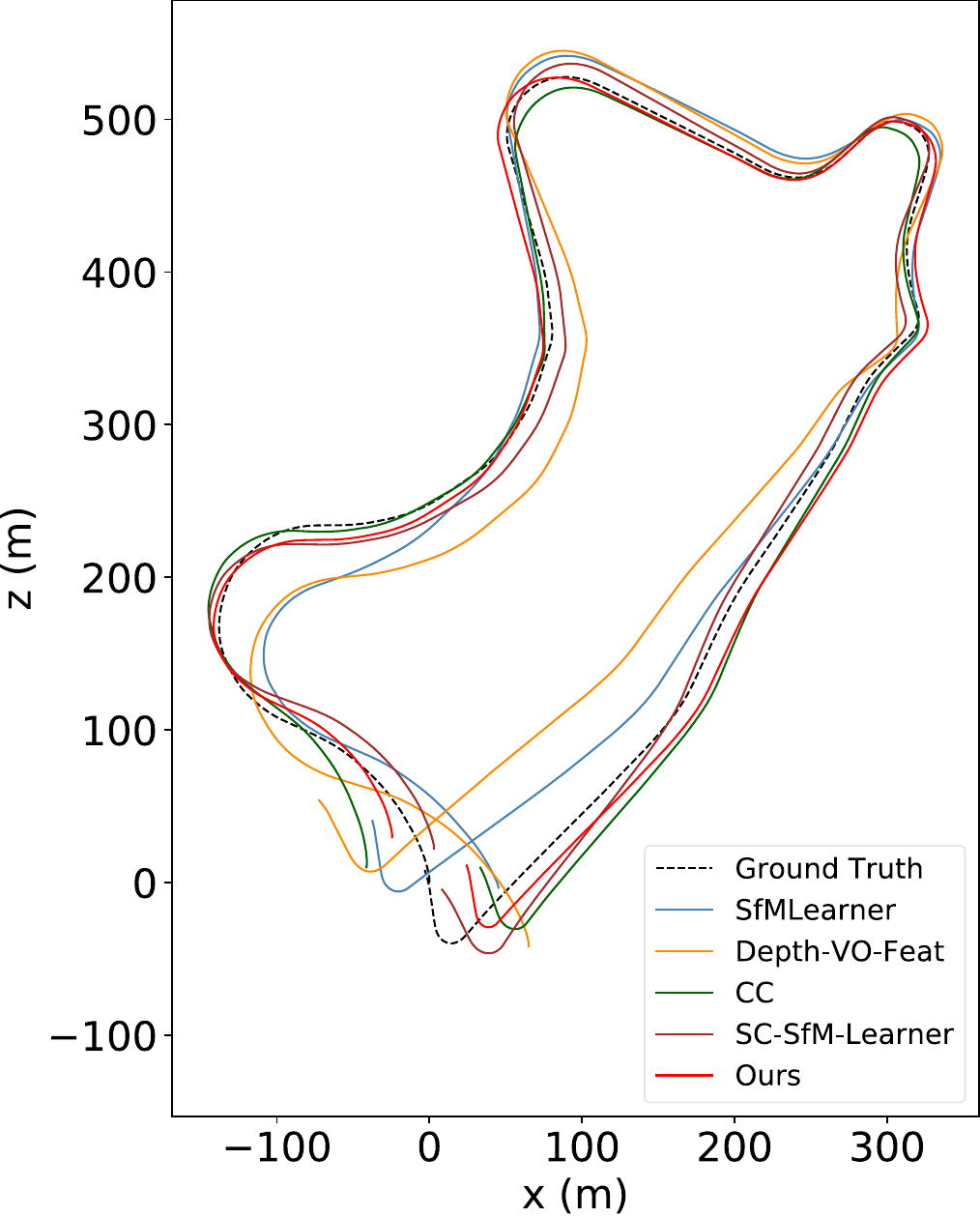}} &
{\includegraphics[width=0.35\linewidth, height=120pt]{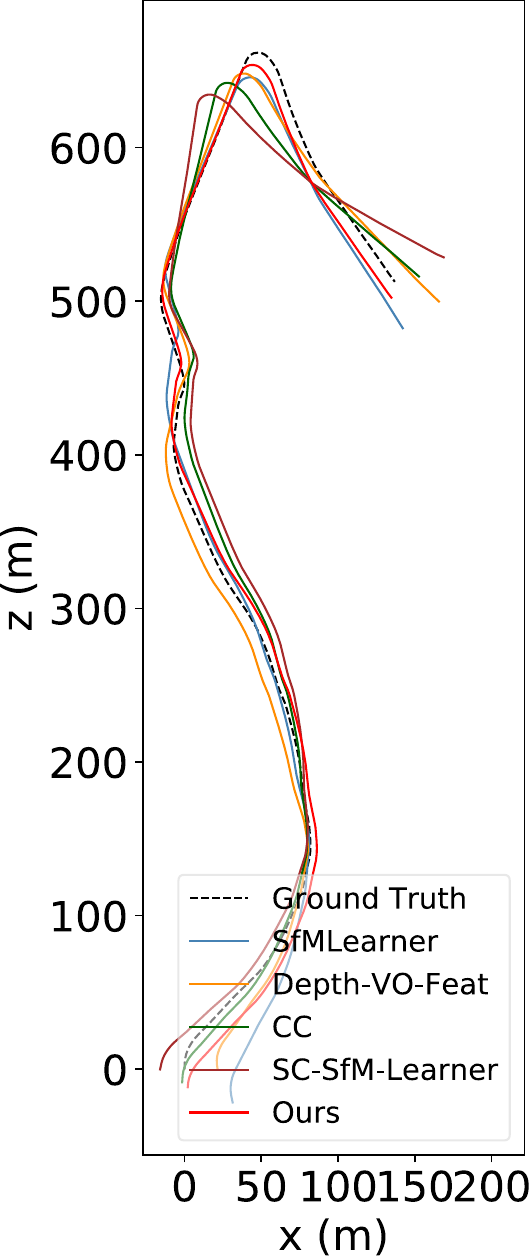}} \\
{\includegraphics[width=0.55\linewidth,height=120pt]{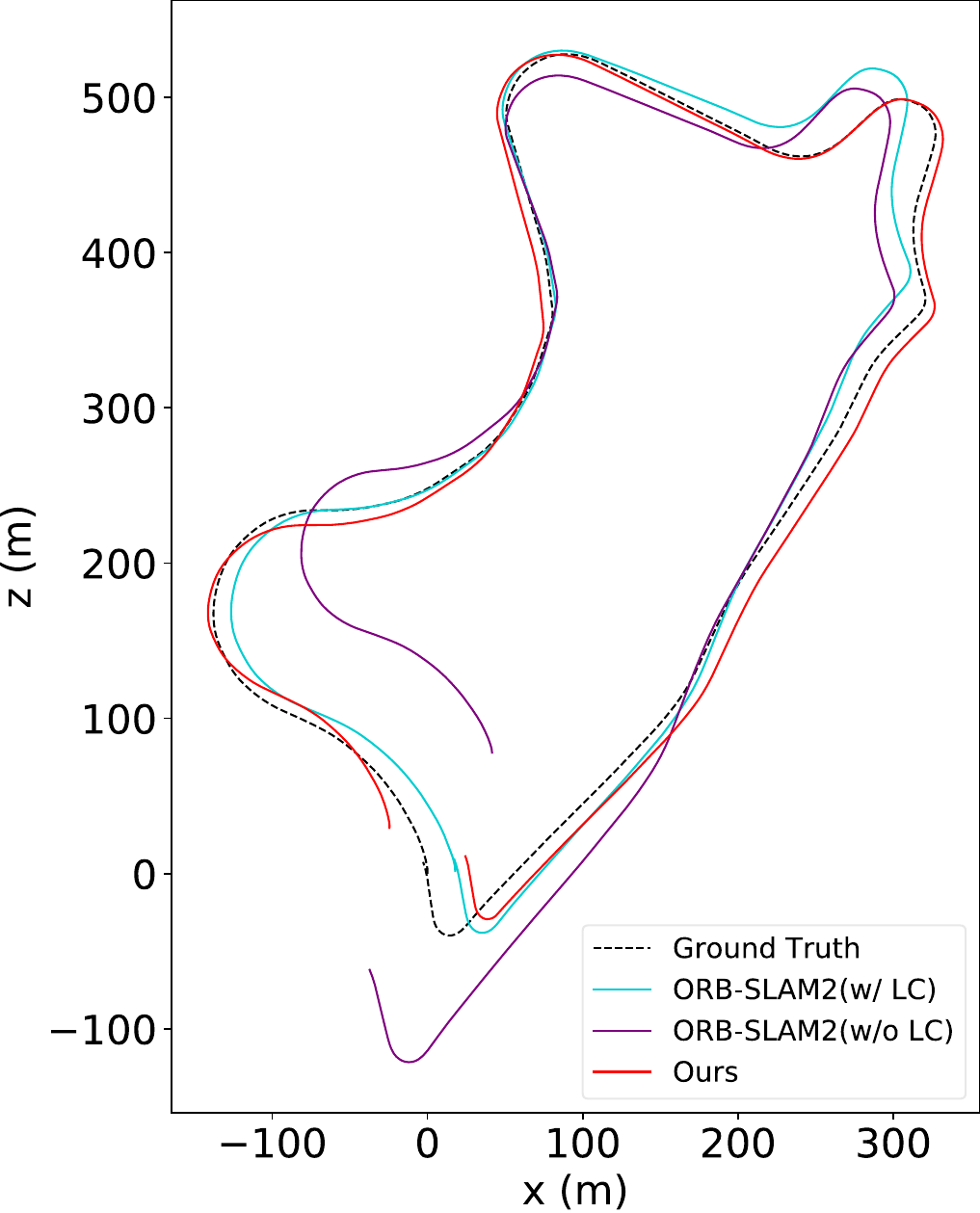}} &
{\includegraphics[width=0.35\linewidth,height=120pt]{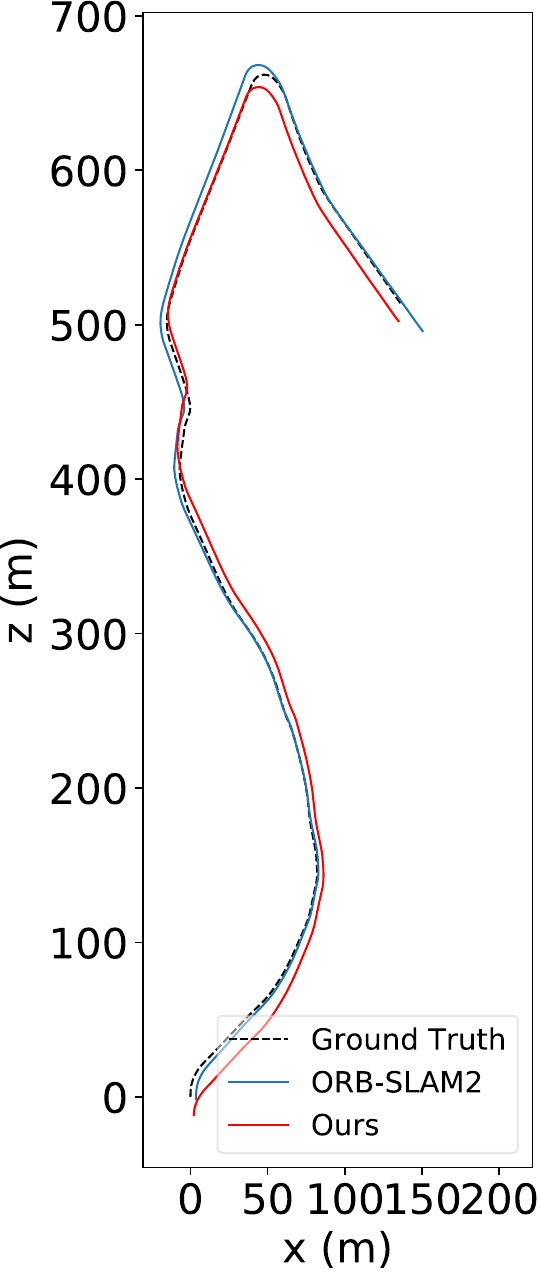}} \\

\end{tabular}
\centering
\caption{Visual odometry results on sequence 09 and 10.}
\label{fig::kitti_odo}
\vspace{-10pt}
\end{figure}

We further extend our system for visual odometry applications. Most of current depth-pose learning methods suffer from error drift when applied on long sequences since the pose network is trained to predict relative pose in short snippets. Recently, Bian \etal \cite{bian2019unsupervised} propose a geometric consistency loss to enforce the long-term consistency of pose prediction and show better results. We test our system with their method and other state-of-the-art depth-pose learning methods on KITTI Odometry datatset. Since monocular systems lack real world scale factor, we align all the predicted trajectory to groundtruth by applying 7DoF (scale + 6DoF) transformation. Table \ref{tab::vo-kitti} shows the results. Because our method essentially mitigates the scale drift of existing depth-pose learning methods with scale inconsistency, we achieve significant performance improvement over state-of-the-art depth-pose learning systems. Although our dense correspondence is learned in an unsupervised manner and no local BA and mapping are used at inference, we achieve comparable results with conventional SLAM systems \cite{mur2017orb}. Figure \ref{fig::kitti_odo} shows the recovered trajectories on two tested sequences respectively. 

\subsection{Generalization on Unseen Ego-motions}
\label{sec::exp_unseen}
To verify the robustness of our method, we design an experiment to test visual odometry application with unseen camera ego-motions. Original sequences in KITTI Odometry dataset are recorded by driving cars with relatively steady velocity, therefore there are nearly no abrupt motions. Meanwhile, the data distributions of relative poses on testing sequences are quite similar to those on training set. We sample the sequences 09 and 10 with different strides to mimic the velocity changes of cameras, and directly test our methods and other depth-pose learning methods, which are all trained on original KITTI Odometry training split, and tested on these new sequences. Table \ref{tab::vo-kitti-sample} shows the results on sequences 09 and 10 which are sampled with stride 3. It is clearly shown that our method is robust and generalize much better on this unseen data distribution, even compared to ORB-SLAM2 \cite{mur2017orb}, which frequently fails and re-initializes under fast motion. More surprisingly, as shown in Figure \ref{fig::firstpage}, all existing depth-pose learning methods relying on 
PoseNet fail to predict reasonable and consistent poses, and produce relatively similar trajectories, which drift far away from the groundtruth trajectory. This might be due to the fact that CNN-based pose estimation acts more like a retrieval method and cannot generalize to unseen data. This interesting finding shares similar spirits with recent works \cite{sattler2019understanding,zhou2019learn}, where the generalization ability of CNN-based absolute pose estimation is studied in depth. With our scale-agnostic system design and the use of conventional two-view geometry, we achieve significantly more robust performance on videos with unseen per-frame ego-motions.

\begin{table}[tb]
\centering
    
    \scalebox{0.75}{
    \setlength\tabcolsep{3pt} 
    \begin{tabular}{l | c c | c c}
     \hline
     Methods & \multicolumn{2}{c|}{Seq. 09} & \multicolumn{2}{c}{Seq. 10} \\
     & $t_{err}$ ($\%$) & $r_{err}$ ($^{\circ}/100m$) & $t_{err}$ ($\%$) & $r_{err}$ ($^{\circ}/100m$) \\
    \hline
    ORB-SLAM2$^\dag$ ~\cite{mur2017orb} & 9.31 & 0.26 & 2.66 & 0.39 \\
    ORB-SLAM2 ~\cite{mur2017orb} & \textbf{2.84} & \textbf{0.25} & \textbf{2.67} & \textbf{0.38} \\
    \hline
    Zhou~\etal~\cite{zhou2017unsupervised} & 11.34 & 4.08 & 15.26 & 4.08 \\
    Deep-VO-Feat \cite{zhan2018unsupervised} & 9.07 & 3.80 & 9.60 & 3.41 \\
    CC \cite{ranjan2019competitive} & 7.71 & 2.32 & 9.87 & 4.47 \\
    SC-SfMLearner \cite{bian2019unsupervised} & 7.60 & 2.19 & 10.77 & 4.63 \\
    Ours & \textbf{6.93}  & \textbf{0.44} & \textbf{4.66} & \textbf{0.62} \\
    \hline
    \end{tabular}
    }
    \vspace{0.08cm}
    \caption{Visual odometry results on KITTI Odometry dataset. The average translation and rotation errors are reported. ORB-SLAM2$^\dag$ indicates that the loop closure is disabled.}
    \vspace{-15pt}
\label{tab::vo-kitti}
\end{table}
\begin{table}[tb]
\centering
    
    \scalebox{0.75}{
    \setlength\tabcolsep{3pt} 
    \begin{tabular}{l | c c | c c}
     \hline
     Methods & \multicolumn{2}{c|}{Seq. 09} & \multicolumn{2}{c}{Seq. 10} \\
     & $t_{err}$ ($\%$) & $r_{err}$ ($^{\circ}/100m$) & $t_{err}$ ($\%$) & $r_{err}$ ($^{\circ}/100m$) \\
    \hline
    ORB-SLAM2 ~\cite{mur2017orb} & X & X & X & X \\
    \hline
    Zhou~\etal~\cite{zhou2017unsupervised} & 49.62 & 13.69 & 33.55 & 16.21 \\
    Deep-VO-Feat \cite{zhan2018unsupervised} & 41.24 & 10.80 & 24.17 & 11.31 \\
    CC \cite{ranjan2019competitive} & 41.99 & 11.47 & 30.08 & 14.68 \\
    SC-SfMLearner \cite{bian2019unsupervised} & 52.05 & 14.39 & 37.22 & 18.91 \\
    Ours & \textbf{7.21}  & \textbf{0.56} & \textbf{11.43} & \textbf{2.57} \\
    \hline
    \end{tabular}
    }
    \vspace{0.08cm}
    \caption{Visual odometry results on KITTI Odometry dataset with large sample stride (stride=3). While ORB-SLAM2 is hard to initialize and keeps losing tracking in this case, our method can produce fairly good prediction. See Figure \ref{fig::firstpage} for plotted trajectories.}
    \vspace{-10pt}
\label{tab::vo-kitti-sample}
\end{table}

\subsection{Generalization on Indoor Datasets}
To further test our generalization ability, we evaluate our method on two indoor datasets: NYUv2 \cite{silberman2012indoor} and TUM-RGBD \cite{sturm2012benchmark} benchmark. Indoor environments are challenging due to the existence of large texture-less regions and much more complex ego-motion (compared to relatively consistent ego-motion on KITTI \cite{geiger2012we}), making the training of most existing self-supervised depth-pose learning method collapse, as shown in Figure \ref{fig::nyuv2}. 
We train our network on NYUv2 raw training set and evaluate the depth prediction on labeled test set. Training images are resized to 192$\times$256 by default. Quantitative results are shown in Table \ref{tab:depth-nyu}. Our method achieves state-of-the-art performance among unsupervised learning baselines. To further study the effects on our system design, we introduce two baseline methods in Table \ref{tab:depth-nyu}: \textit{PoseNet} baseline is built by substituting our optical flow and two-view triangulation module with a PoseNet-like architecture, where relative pose is directly predicted with a convolutional neural network, and \textit{PoseNet-Flow} baseline uses optical flow as input for PoseNet branch to predict relative pose. See supplementary material for more details about these two baselines. Our proposed system achieves a large performance gain, indicating the effectiveness and robustness of our system design.
\begin{table}[t]
\small
\setlength\tabcolsep{4.5pt} 
\begin{center}
\centering
\scalebox{0.9}{

\begin{tabular}{lccc|ccc}
& \multicolumn{3}{c|}{Error} & \multicolumn{3}{c}{Accuracy, $\delta$ } \\ \cline{2-4}  \cline{5-7}
Method  & rel & log10 & rms &$<$$1.25$&$<$$1.25^2$& $<$$1.25^3$\\ 
\hline
{ Make3D ~\cite{saxena2008make3d}}  & 0.349 & - & 1.214 & 0.447 & 0.745 & 0.897 \\
{ Li \etal ~\cite{li2015depth}}  & 0.232 & 0.094 & 0.821 & 0.621 & 0.886 & 0.968 \\
{ MS-CRF \cite{xu2017multi}}  & 0.121 & 0.052 & 0.586 & 0.811 & 0.954 & 0.987 \\ 
{ DORN \cite{fu2018deep}}  & 0.115 & 0.051 & 0.509 & 0.828 & 0.965 & 0.992 \\ 
\hline
{Zhou \etal \cite{zhou2019moving}}  & 0.208 & 0.086 & 0.712 & 0.674 & 0.900 & 0.968 \\
{PoseNet} & 0.283 & 0.122 & 0.867 & 0.567 & 0.818 & 0.912 \\
{PoseNet-Flow} & 0.221 & 0.091 & 0.764 & 0.659 & 0.883 & 0.959 \\
{ Ours }  & 0.201 & 0.085 & 0.708 &
0.687 & 0.903 & 0.968 \\
{ Ours (448$\times$576)}  & \textbf{0.189} & \textbf{0.079} &  \textbf{0.686} &
\textbf{0.701} & \textbf{0.912} & \textbf{0.978} \\
\hline
\end{tabular}}
\vspace{0.01cm}
\caption{Results on NYUv2 depth estimation. Supervised methods are shown in the first rows. \textit{PoseNet} indicates replacing flow and triangulation module with PoseNet in our system. \textit{PoseNet-Flow} indicates using optical flow as input for PoseNet.}
\label{tab:depth-nyu}
\end{center}
\vspace{-15pt}
\end{table}
\begin{figure}[tb]
\scriptsize
\setlength\tabcolsep{3.0pt} 

\scalebox{0.95}{
\begin{tabular}{cccc}
{\includegraphics[width=0.23\linewidth]{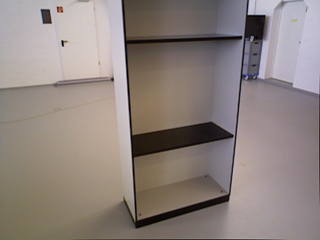}} &
{\includegraphics[width=0.23\linewidth]{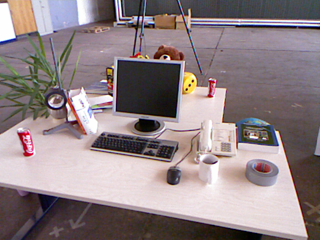}} &
{\includegraphics[width=0.23\linewidth]{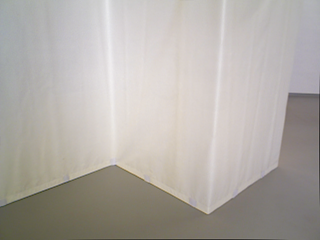}} &
{\includegraphics[width=0.23\linewidth]{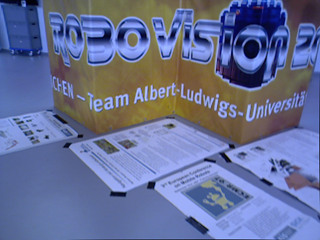}} \\
\end{tabular}}
\begin{tabular}{cccc}
{\includegraphics[width=0.23\linewidth]{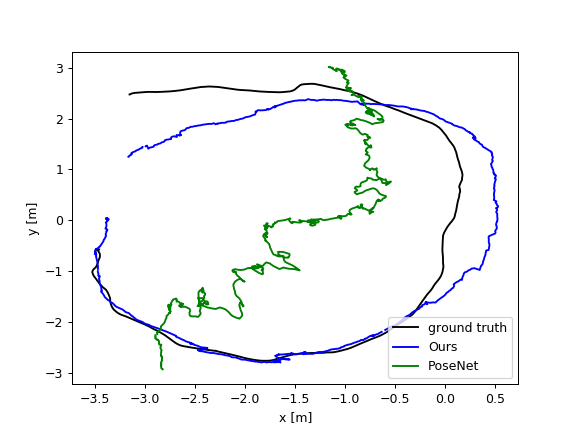}} &
{\includegraphics[width=0.23\linewidth]{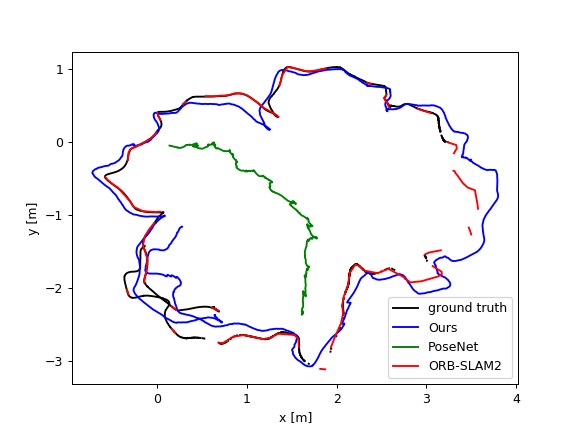}} &
{\hspace{-1em}\includegraphics[width=0.23\linewidth]{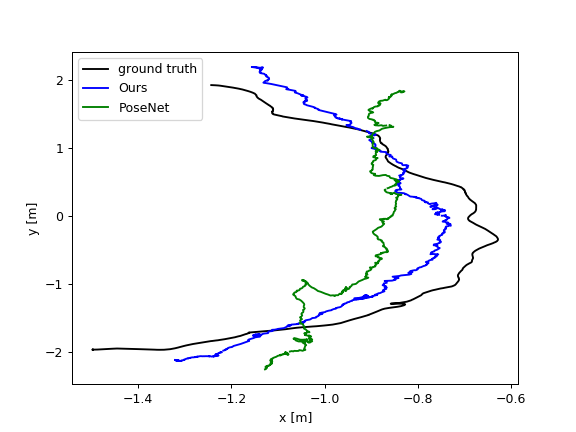}} &
{\hspace{-1em}\includegraphics[width=0.23\linewidth]{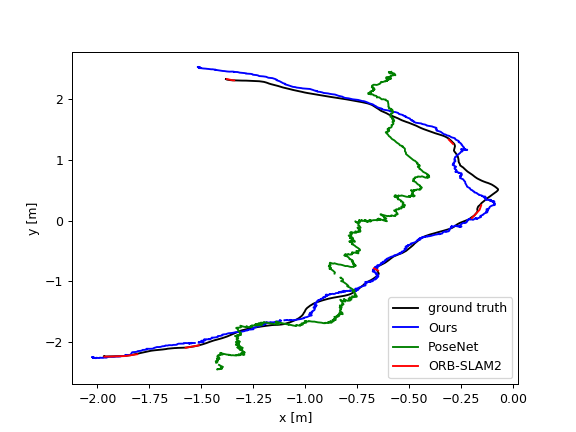}}\\
\end{tabular}
\centering
\caption{Visual odometry results on TUM RGBD dataset. Our proposed system can still work well with large textureless regions (the 1st and the 3rd cases), complex camera motions (the 2nd case) and different lighting conditions (the 4th case), demonstrating improved robustness compared to the baseline. \textbf{Better viewed when zoomed in.}}
\label{fig::tum}
\vspace{-10pt}
\end{figure}

In addition, we test our method on TUM-RGBD \cite{sturm2012benchmark} dataset, which is widely used for evaluating visual odometry and SLAM systems \cite{mur2017orb, whelan2015elasticfusion}. This dataset is collected mainly by hand-held cameras in indoor environments, and consists of various challenging conditions such as extreme textureless regions, moving objects, and abrupt motions, etc. We follow the same train/test setting as \cite{xue2019beyond}. Figure \ref{fig::tum} shows four trajectory results. The PoseNet-like baseline fails to generalize under this setting and produce poor results. Conventional SLAM system like ORB-SLAM2 works well if there exists rich textures but tends to fail when large textureless region occurs, such as the first and the third cases shown in Figure \ref{fig::tum}. In most cases, thanks to joint dense correspondence learning, our method can establish accurate pixel associations to recover camera ego-motions and produce reasonably well trajectories, again demonstrating our improved generalization.

\begin{figure}[tb]
\scriptsize
\setlength\tabcolsep{1.0pt} 
\renewcommand{\arraystretch}{1.0}

\begin{tabular}{ccc}
{\includegraphics[width=0.32\linewidth]{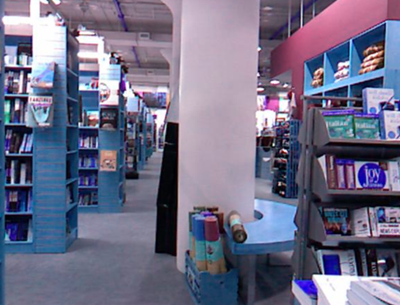}} &
{\includegraphics[width=0.32\linewidth]{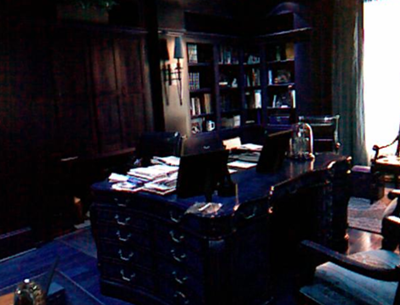}} &
{\includegraphics[width=0.32\linewidth]{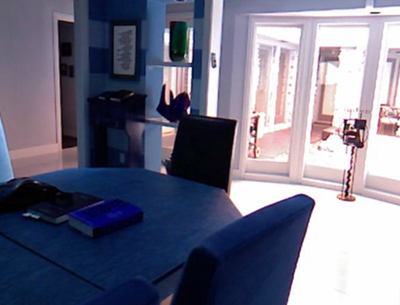}} \\

{\includegraphics[width=0.32\linewidth]{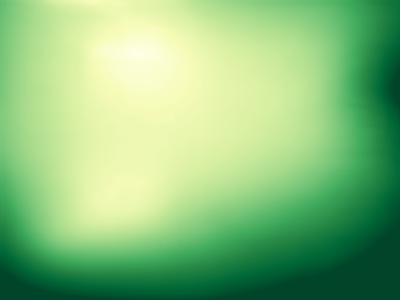}} &
{\includegraphics[width=0.32\linewidth]{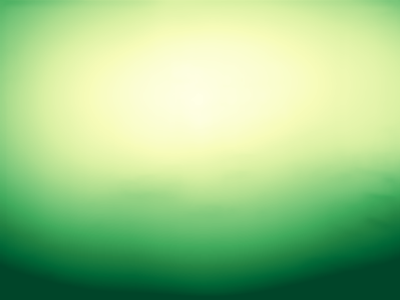}} &
{\includegraphics[width=0.32\linewidth]{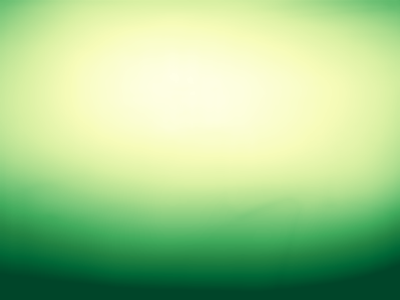}} \\
{\includegraphics[width=0.32\linewidth]{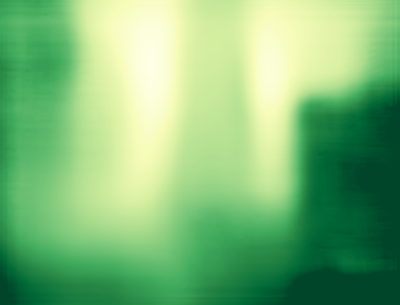}} &
{\includegraphics[width=0.32\linewidth]{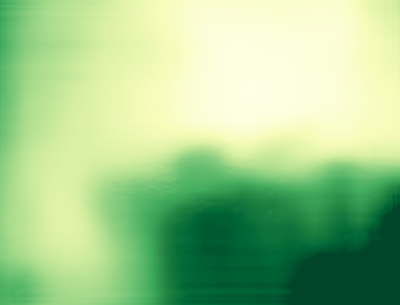}} &
{\includegraphics[width=0.32\linewidth]{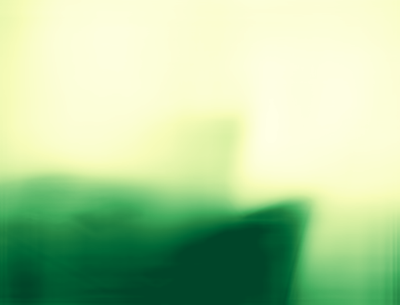}}\\

{\includegraphics[width=0.32\linewidth]{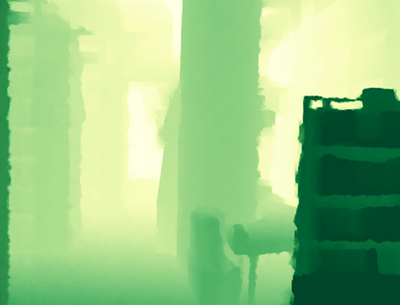}} &
{\includegraphics[width=0.32\linewidth]{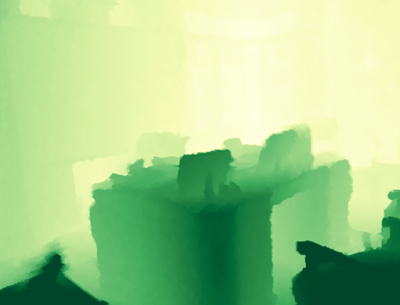}} &
{\includegraphics[width=0.32\linewidth]{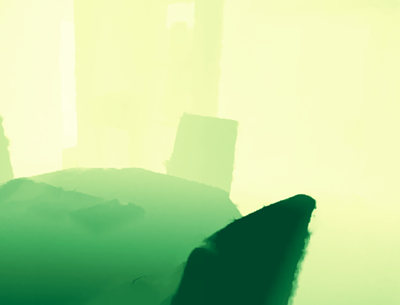}} \\
\end{tabular}
\centering
\caption{Depth estimation results on NYUv2 test data. \textbf{Top to bottom}: Input image, PoseNet baseline prediction, our prediction and depth groundtruth. PoseNet baseline fails to generalize for this indoor environment, which is also reported in \cite{zhou2019moving}.}
\label{fig::nyuv2}
\vspace{-10pt}
\end{figure}

\subsection{Discussion}
Our experiments show that in addition to that our method maintains on par or even better performance on the widely tested KITTI benchmark, we achieve significant improvement on robustness and generalization from a variety of different aspects. This gain on generalization comes from our two novel designs as follows: 1) direct camera ego-motion prediction from optical flow, and 2) explicit scale alignment between the depth and the triangulated 3D structure. 
Our findings suggest that optical flow, which does not suffer from scale ambiguity naturally, is a more robust visual clue compared to relative pose estimation for deep learning models, especially under challenging scenarios. Likewise, explicitly handling the scale of depth and pose is still crucial for deep learning based visual SLAM. 
However, our current system cannot handle multi-view images where the motion magnitude is beyond the cost volume of optical flow, and pure rotation cannot be handled online with the two-view triangulation module. 


\section{Conclusion}
In this paper, we propose a novel system which tackles the scale inconsistency for self-supervised joint depth-pose learning, by (1) directly recovering relative pose from optical flow and (2) explicit scale alignment between depth and pose via triangulation. Experiments demonstrate that our method achieves significant improvement on both accuracy and generalization ability over existing methods. 
Handling the above mentioned failure cases, developing general correspondence prediction and integration with back-end optimization could be interesting future directions.


\section*{Acknowledgements}
This work was partially supported by NSFC (61725204, 61521002), BNRist and MOE-Key Laboratory of Pervasive Computing.

\clearpage
{\small
\bibliographystyle{ieee_fullname}
\bibliography{egbib}
}

\clearpage
\section*{Appendix}
\appendix
This document provides a list of supplemental materials that accompany the main paper.
\begin{itemize}
    \item \textbf{Discussion on Scale-Invariant Design} - We provide more detailed discussion for the scale-invariant design in our system in Section~\ref{sec:scale-invariant}.
    \item \textbf{Derivation of Triangulation Module} - We include the detailed derivation of differentiable triangulation module in Section~\ref{sec:tri}.
    \item \textbf{Details for PoseNet and PoseNet-Flow} - We introduce more details and results about PoseNet and PoseNet-Flow in Section~\ref{sec:flowposenet}. 
    \item \textbf{Additional Results and Discussion for PoseNet-Flow} - We present additional experiemental results for PoseNet-Flow on visual odometry in Section~\ref{sec:exp-posenetflow}.
    \item \textbf{Implementation Details} - We provide more implementation details about network architectures and system hyperparameters in Section~\ref{sec:details}.
    \item \textbf{Additional Comparison on sampled KITTI Odometry dataset} - We show more comparsion results about sampled KITTI Odometry dataset in Section~\ref{sec:odo}.
    \item \textbf{Numerical Results of TUM-RGBD dataset} - We report quantitative results for TUM-RGBD dataset in Section~\ref{sec:tum}.
    \item \textbf{Additional Visualizations} - In Section~\ref{sec:vis}, we provide additional visualizations generated by our system on different datasets.
\end{itemize}

\section{Discussion on Scale-Invariant Design}
\label{sec:scale-invariant}
Given a pair of input images, assume that the fundamental matrix can be accurately recovered from point correspondence and no additional priors exist, the relative translation of the pair should be up to an arbitrary scale. On the other hand, the monocular depth estimation aims to use learned priors from data to directly infer the corresponding depth image. Assume that the intrinsic parameters of the camera are known and fixed, the system can possibly make use of the common priors such as the height of human, the width of the car as well as subtle structural clues to infer the monocular depth, which does not suffer from the scale ambiguity problem. 

Most previous works (e.g., \cite{zhou2017unsupervised}) use two separate convolutional neural networks to learn both monocular depth and relative pose, and directly put photometric consistency constraint by using the predicted relative pose to reproject the predicted depth. This makes the assumption that the scale of the predicted relative pose should correspond to the predicted monocular depth, which means that the relative pose estimation is required to not only learn the feature matching and relative pose recovery, but also implicitly learn the scale priors which are exactly the same as the monocular depth estimation is required to learn. This requires the network to firstly infer scale from two input images respectively, and implicitly integrate the predicted scale into the recovered relative pose, making the learning of pose prediction network extremely hard and degrade its generalization capability.

Our method explicitly resolves this problem with two novel designs:
\begin{itemize}
    \item \rom{1}. We use an optical flow network to specifically learn pixelwise matching, then solve the fundamental matrix and recover the relative pose up an arbitrary scale.
    \item \rom{2}. We triangulate the predicted correspondence and explicitly align the predicted depth to the triangulated point clouds to compute the error map. 
\end{itemize}

In this way, the relative pose prediction is not required to implicitly learn the scale priors. This significantly improves the generalization both for training on indoor environments and inference on video sequences with unseen camera ego-motion. Note that, the two designs are necessary to be coupled together. Suppose that if the system only employs design \rom{1} without aligning the depth to the triangulated point clouds, the joint training cannot converge because it is impossible to fit the scale of the depth estimation network to an arbitrary scale of relative pose. 

Based on the previous discussion, we can infer that our system is robust under the circumstances where the camera intrinsic parameters are known and fixed. When the camera intrinsic parameters are flexible across different sequences on training and inference, only under the assumption that the monocular depth estimation network can automatically learn the camera calibration from structural clues in the single image can our method still accurately recover the depth image. Otherwise, further system designs on the monocular depth network are required to disentangle the influence of different camera field of view to make the learning problem feasible.

\section{Derivation of Triangulation Module}
\label{sec:tri}
We adopt mid-point triangulation method to build an up-to-scale 3D structure from 2D correspondences and relative pose. Mid-point triangulation problem could be easily solved with linear algorithms. The objective function is as follows:
\begin{equation}
\vec x^* = \underset{\vec x}{\operatorname{argmin}}{\ \varphi} = \underset{\vec x}{\operatorname{argmin}}{\ [d(\vec L_1, \vec x)]^2 + [d(\vec L_2, \vec x)]^2}
\label{eq::mid-point1}
\end{equation}

\noindent Where $\vec L_1 = \{\vec p=\vec c_1+{\lambda}_1 \vec n_1\mid\lambda_1 \in \mathbb{R}\}$ and $\vec L_2 = \{\vec p=\vec c_2+{\lambda}_2\vec n_2\mid\lambda_2 \in \mathbb{R}\}$ are two camera rays generated with optical flow correspondence, and $d$ denotes the euclidean distance. $\vec c_i = -R_i^T\vec t_i $ is the ray origin, where $[R,\vec t]$ is the camera extrinsic, and $\vec n_i = R_i^T K^{-1}{[x_0, y_0, 1]}^T $ is the ray direction, where $[x_0,y_0]$ is the pixel coordinate. The objective function could be written as:
\begin{equation}
    \varphi(\vec x, \lambda_1, \lambda_2) = {\Vert \vec c_1+{\lambda}_1\vec n_1 - \vec x\Vert}^2 + {\Vert \vec c_2+{\lambda}_2\vec n_2 - \vec x\Vert}^2
\label{eq::mid-point2}
\end{equation}

To minimize $\varphi(\vec x)$, we need $\frac{\partial\varphi}{\partial \vec x} = 0$ which easily gives us:
\begin{equation}
\vec x = \frac{(\vec c_1+{\lambda}_1\vec n_1) + (\vec c_2+{\lambda}_2\vec n_2)}{2}
\label{eq:x}
\end{equation}
After substitution of $\vec x$, the cost function becomes:
\begin{equation}
    \varphi(\vec x, \lambda_1, \lambda_2) = \frac{1}{2}{\Vert (\vec c_1+{\lambda}_1\vec n_1) - (\vec c_2+{\lambda}_2\vec n_2)\Vert}^2
\label{eq::mid-point3}
\end{equation}
Then we have:
\begin{equation}
\begin{aligned}
    \frac{\partial \varphi}{\partial \lambda_1} &= \vec n_1^T(\lambda_1 \vec n_1-\lambda_2 \vec n_2+\vec c_1-\vec c_2) = 0 \\
    \frac{\partial \varphi}{\partial \lambda_2} &= \vec n_2^T(\lambda_2 \vec n_2-\lambda_1 \vec n_1+\vec c_2-\vec c_1) = 0
\label{eq::mid-point4}
\end{aligned}
\end{equation}
From these two linear equations, the solutions of $\lambda_1$ and $\lambda_2$ could be expressed as:

\begin{equation}
    \left[\begin{array}{c}
         \lambda_1 \\
         \lambda_2
    \end{array} \right] = A\left[\begin{array}{cc}
    {\Vert \vec n_2\Vert}^2 & \vec n_1^T \vec n_2 \\
    \vec n_2^T \vec n_1 & {\Vert \vec n_1\Vert}^2
    \end{array}
    \right] \left[\begin{array}{c}
    \vec n_1^T(\vec c_2-\vec c_1) \\
    \vec n_2^T(\vec c_1-\vec c_2)
    \end{array}
    \right]
\label{eq::mid-point5}
\end{equation}

\begin{equation}
A = \frac{1}{{\Vert \vec n_1\Vert}^2{\Vert \vec n_2\Vert}^2 - (\vec n_1^T\vec n_2)^2}
\end{equation}
The triangulation solution $\vec x$ is then computed with Eq. (\ref{eq:x}). By this way, the triangulation module is naturally differentiable.

\section{Details for PoseNet and PoseNet-Flow}
\label{sec:flowposenet}
We implement two baseline methods, named \textit{PoseNet} and \textit{PoseNet-Flow}, to compare with our method. \textit{PoseNet} system takes image pairs as input, predicts monocular depth and relative pose by depth and pose branch, respectively. The depth branch uses the same network as our system and the pose branch adopts standard PoseNet~\cite{kendall2015posenet}. Following previous PoseNet-based unsupervised depth pose joint learning methods~\cite{zhou2017unsupervised, bian2019unsupervised}, we utilize photometric loss and depth reprojection loss to train the network. For \textit{PoseNet-Flow} system, we add a flow network to generate optical flow, and feed generated optical flow, rather than RGB image pair, to PoseNet for relative pose estimation. The flow network is the same as that of our system. The depth network and the depth-pose training objectives remain the same as \textit{PoseNet} system. We adopt two-stage training stragegy for \textit{PoseNet-Flow} system. In the first stage we train the optical flow network. Then the flow network is frozen and both the depth and pose networks are joint trained.

\section{Additional Results and Discussion for PoseNet-Flow}
\begin{figure}[tb]
\scriptsize
\begin{tabular}{cc}
{\includegraphics[width=0.55\linewidth,height=120pt]{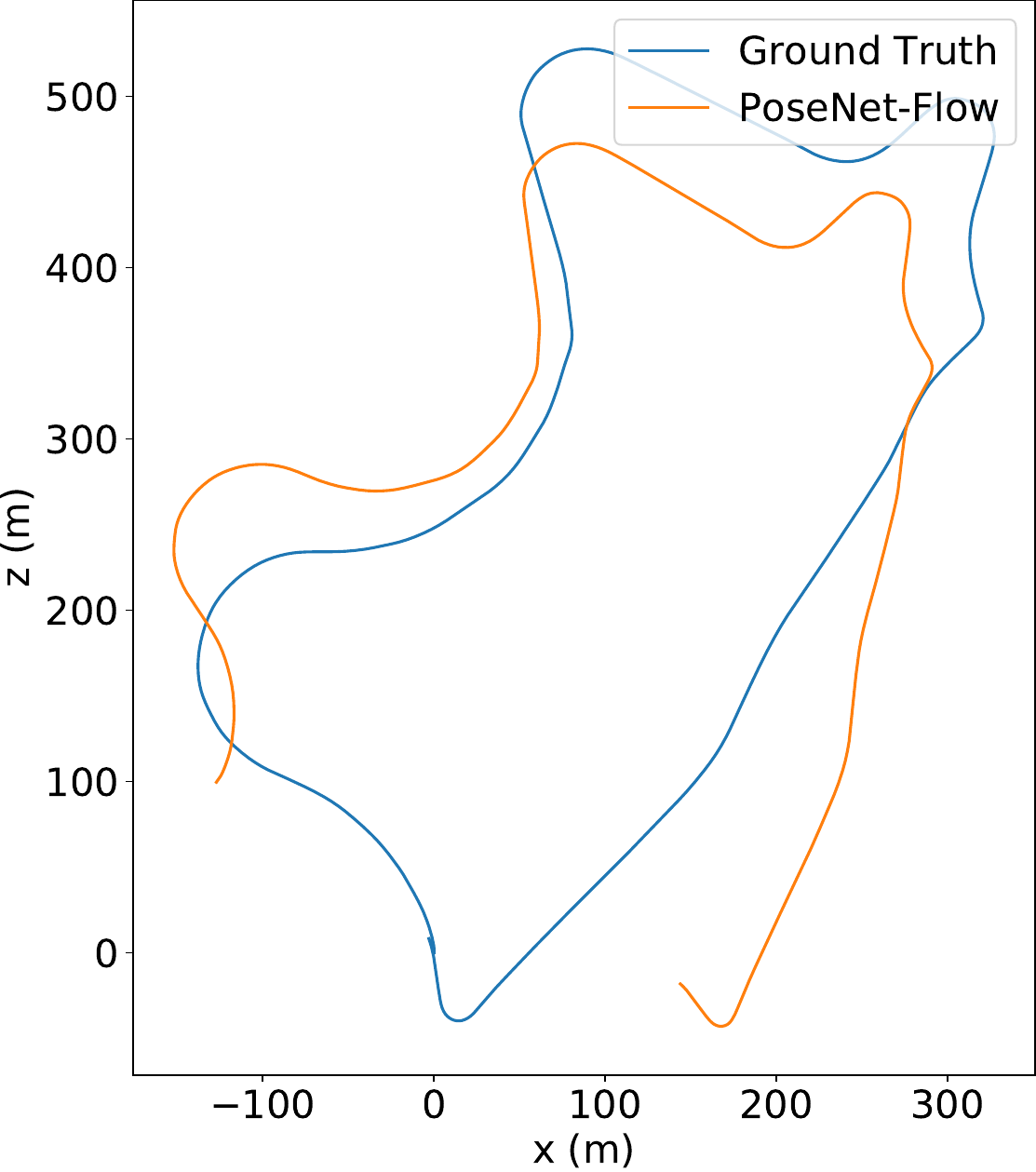}} &
{\includegraphics[width=0.35\linewidth, height=120pt]{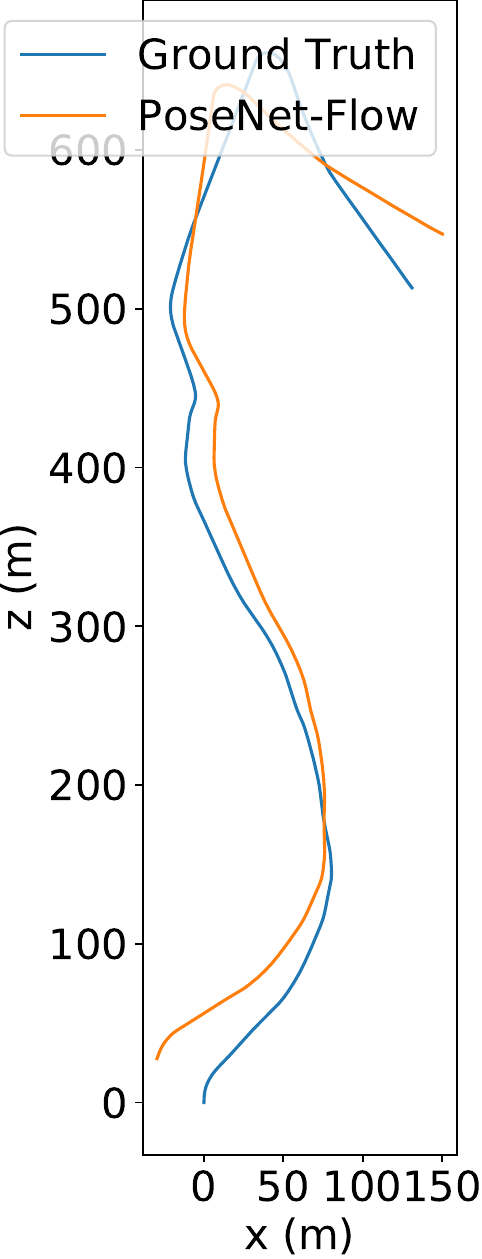}} \\

\end{tabular}
\centering
\caption{Visual odometry results of PoseNet-Flow method on original sequence 09 and 10.}
\label{fig:supp_flowodo}
\vspace{0pt}
\end{figure}

\begin{figure}[tb]
\scriptsize
\begin{tabular}{cc}
{\includegraphics[width=0.55\linewidth,height=120pt]{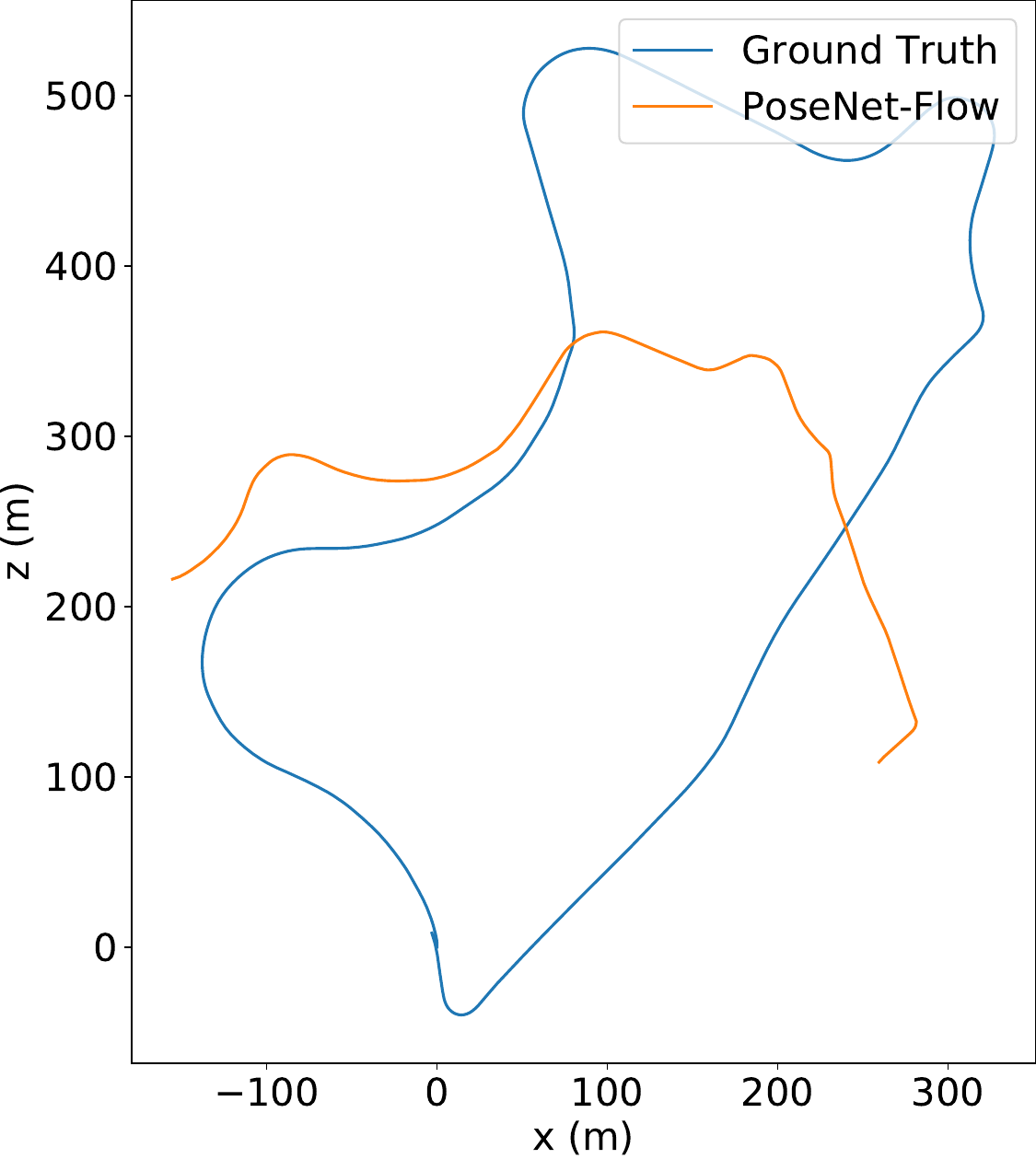}} &
{\includegraphics[width=0.35\linewidth, height=120pt]{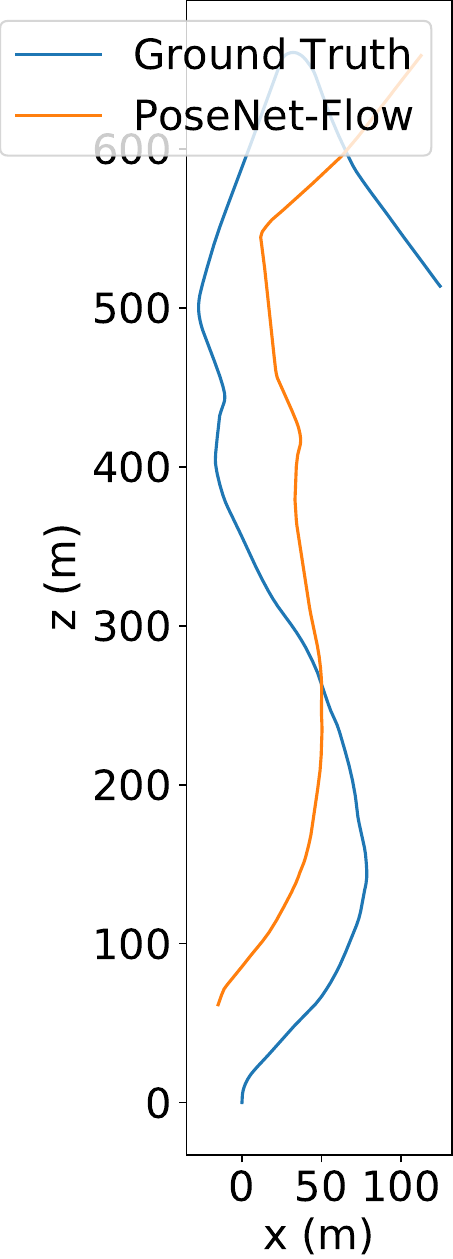}} \\

\end{tabular}
\centering
\caption{Visual odometry results of PoseNet-Flow method on sampled sequence 09 and 10 with stride 3.}
\label{fig:supp_flowodo_s3}
\vspace{0pt}
\end{figure}
\label{sec:exp-posenetflow}
Table \ref{tab:depth-nyu} shows the depth estimation results of \textit{PoseNet} and \textit{PoseNet-Flow} in indoor NYUv2 dataset. Due to complex camera motions and large textureless regions, traditional \textit{PoseNet} method fails to generate plausible predictions. \textit{PoseNet-Flow} uses optical flow for pose regression, thus improves the interpretability of the system and makes learning problem easier. This is also discussed in \cite{zhou2019moving}. To further explore the capacity of \textit{PoseNet-Flow} system, we conduct experiments on KITTI Odometry dataset. We use two consecutive images as training pairs. Figure \ref{fig:supp_flowodo} and Figure \ref{fig:supp_flowodo_s3} show the results of standard KITTI dataset and sampled KITTI dataset with stride 3. While the \textit{PoseNet-Flow} system could produce feasible results on NYUv2 and standard KITTI dataset, it still tends to fail on unseen ego-motions. This could possibly due to the nature of trained PoseNet that it performs more like image retrieval rather than solving physical constraints and thus works well only on the test data which is similar with training samples. On the contrary, our method works well under all these challenging scenarios, showing much improved robustness and generalization ability.

\section{Implementation Details}
\label{sec:details}

Here we introduce more details about network architectures and training objectives used in our system. 

For depth estimation network, we adopt a same encoder-decoder network with skip connections as proposed in \cite{godard2018digging}. Specifically, ResNet-18 is used as encoder and DispNet \cite{mayer2016large, godard2017unsupervised} is used as decoder with ELU nonlinearities for all conv layers except output layer, where we use sigmoids and convert the output disparity to depth with $D=1/(ad+b)$. $a$ and $b$ are set to be 0.1 and 100 to constrain the range of output depth. We only supervise the largest scale of depth output, and replace the nearest upsampling layers in decoder with bilinear upsampling, which makes the training more stable. The depth loss consists of three parts, triangulation depth loss $L_d$, reprojection loss $L_p$ and edge-aware depth smoothness loss $L_s$. The detailed descriptions of $L_d$ and $L_p$ are included in the main paper. Given image input $I_t$ and disparity prediction $d_t$, depth smooth loss $L_s$ is computed as follows:
\begin{equation}
L_s = \lvert \partial_x d_t^{n}\rvert\ e^{-\lvert \partial_x I_t \rvert} + \lvert \partial_y d_t^{n}\rvert\ e^{-\lvert \partial_y I_t \rvert}
\label{eq:depth_smooth}
\end{equation}

\noindent where $d_t^{n} = d_t / \overline{d_t}$ is the normalized disparity prediction to avoid depth shrinking, proposed by \cite{wang2018learning}.

For flow estimation network, we adopt the PWCNet \cite{sun2018pwc} as backbone for predicting forward and backward optical flow of an image pair. We utilize the backward warping method proposed in \cite{wang2018occlusion} to explicitly handle occlusions. Generated occlusion masks are not only used as a better supervision for the optical flow, but also for sampling reliable pixel matches when solving relative pose and triangulation. Optical flow is predicted and supervised at three different scales. Following \cite{yin2018geonet, zou2018df}, we use a 
combination of L1 loss, SSIM loss \cite{wang2004image} and flow smoothness loss for flow supervision. Therefore, the total flow loss $L_f$ is expressed as:
\begin{equation}
L_f = (1-\alpha) \Vert I_a - I_b \Vert + \frac{\alpha}{2}(1-SSIM(I_a, I_b)) + \beta L_{fs}
\label{eq:flow}
\end{equation}
\noindent where $L_{fs}$ is the flow smoothness loss which has a similar formulation as Eq. (\ref{eq:depth_smooth}). $\alpha$ and $\beta$ are set to be 0.85 and 0.1 respectively.

\begin{figure}[tb]
\scriptsize
\setlength\tabcolsep{2pt} 
\begin{tabular}{cc}
Image & Epipolar Lines \\
\includegraphics[width=0.45\linewidth]{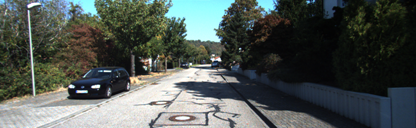} & 
\includegraphics[width=0.45\linewidth]{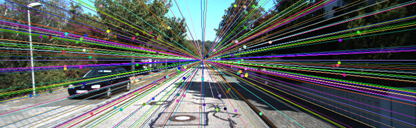} \\
Dense Triangulation & Angle Mask \\
\includegraphics[width=0.45\linewidth]{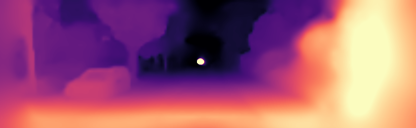} &
\includegraphics[width=0.45\linewidth]{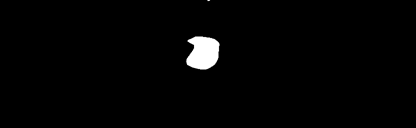} \\

\end{tabular}
\centering
\caption{The white area in angle mask means extremely small angles between two rays or negative triangulation depths. Small ray angles and negative depths often happen near epipoles, which are the intersection points of all epipolar lines. }
\label{fig:supp_angle}
\vspace{0pt}
\end{figure}
\begin{figure}[tb]
\scriptsize
\setlength\tabcolsep{2pt} 
\begin{tabular}{cc}
\includegraphics[width=0.45\linewidth]{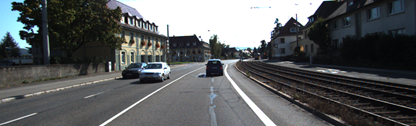} & 
\includegraphics[width=0.45\linewidth]{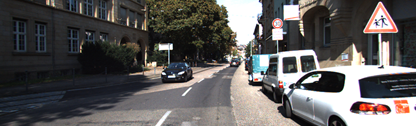} \\
\includegraphics[width=0.45\linewidth]{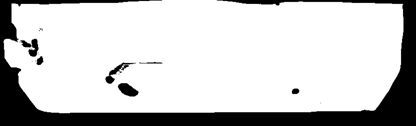} &
\includegraphics[width=0.45\linewidth]{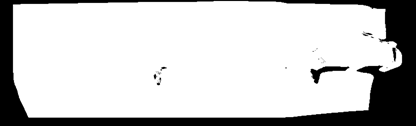} \\
\includegraphics[width=0.45\linewidth]{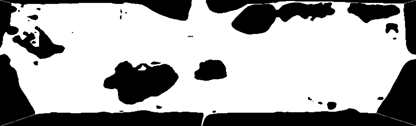} &
\includegraphics[width=0.45\linewidth]{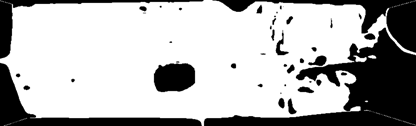} \\

\end{tabular}
\centering
\caption{\textbf{Top to bottom:} Image, occlusion mask, inlier map. The inlier map is converted to binary mask for better visualization. The occlusion masks and inlier maps could successfully filter out occlusions and non-rigid regions respectively.}
\label{fig:supp_inlier}
\vspace{0pt}
\end{figure}
For relative pose estimation, we recover it by solving fundamental matrix. Specifically, we first compute optical flow forward-backward distance map $D_{fb}$ by flow warping. Then forward-backward score map $M_s$ is generated as $M_s = 1/(0.1+D_{fb})$. Together, $M_o*M_s$ is used for sampling accurate correspondences from dense flow. We sample the top 20\% correspondences according to score map and then randomly sample 6k matches. We perform this sampling strategy, rather than directly top sampling, to discourage spatial accumulation of sampled matches. Then we run the normalized 8-point algorithm in RANSAC loop to solve fundamental matrix. The RANSAC inlier threshold and desirable confidence are set to be 0.1 and 0.99 respectively. After solving fundamental matrix, we decompose it into $[R,t]$ and further triangulate matches for all four $[R,t]$ solutions. We choose the one which has the most triangulated points in front of both cameras as final relative pose. An inlier score map $M_r$ is generated from fundamental matrix to mask out non-rigid regions such as moving objects and bad matches. See examples in Figure \ref{fig:supp_inlier}. Specifically, we compute the distance from one pixel to its corresponding epipolar line, resulting in distance map $D_{epi}$. The inlier score map is computed as $M_r = (D_{epi} < 0.5) / (1.0 + D_{epi})$. Again we perform top score sampling and random sampling from $M_r*M_s*M_o$ to acquire 6k matches. We filter out the matches which have extremely small ray angles or have invalid reprojection. To be specific, given two camera rays $\vec L_1 = \{\vec p=\vec c_1+{\lambda}_1 \vec n_1\mid\lambda_1 \in \mathbb{R}\}$ and $\vec L_2 = \{\vec p=\vec c_2+{\lambda}_2 \vec n_2\mid\lambda_2 \in \mathbb{R}\}$, where $\vec c_i$ is the ray origin and $\vec n_i$ is the ray direction, we could have $\vec v = \vec c_2 + \left< \vec c_1 - \vec c_2, \vec n_2 \right> \vec n_2 - \vec c_1$. Then the cosine value of angle between $\vec v$ and $\vec n_1$ is computed. We filter out the regions where the cosine value is smaller than 0.001. See an example in Figure \ref{fig:supp_angle}. After filtering, matches are further triangulated to 3D structure, and then used for scale alignment and supervision of depth prediction. 

\begin{table}[tb]
\centering
    
    \scalebox{0.75}{
    \setlength\tabcolsep{3pt} 
    \begin{tabular}{l | c c | c c}
     \hline
     Methods & \multicolumn{2}{c|}{Seq. 09} & \multicolumn{2}{c}{Seq. 10} \\
     & $t_{err}$ ($\%$) & $r_{err}$ ($^{\circ}/100m$) & $t_{err}$ ($\%$) & $r_{err}$ ($^{\circ}/100m$) \\
    \hline
    ORB-SLAM2$^\dag$ ~\cite{mur2017orb} & 11.12 & \textbf{0.33} & 2.97 & 0.36 \\
    ORB-SLAM2 ~\cite{mur2017orb} & \textbf{2.37} & 0.40 & \textbf{2.97} & \textbf{0.36} \\
    \hline
    Zhou~\etal~\cite{zhou2017unsupervised} & 24.75 & 7.79 & 25.09 & 11.39 \\
    Depth-VO-Feat \cite{zhan2018unsupervised} & 20.54 & 6.33 & 16.81 & 7.59 \\
    CC \cite{ranjan2019competitive} & 24.49 & 6.58 & 19.49 & 10.13 \\
    SC-SfMLearner \cite{bian2019unsupervised} & 33.35 & 8.21 & 27.21 & 14.04 \\
    Ours & \textbf{7.02}  & \textbf{0.45} & \textbf{4.94} & \textbf{0.64} \\
    \hline
    \end{tabular}
    }
    \vspace{0.08cm}
    \caption{Visual odometry results on sampled sequence 09 and 10 with stride 2. The average translation and rotation errors are reported. ORB-SLAM2$^\dag$ indicates disablement of loop closure.}
    \vspace{-5pt}
\label{tab::vo-kitti-s2}
\end{table}

\begin{table}[tb]
\centering
    
    \scalebox{0.75}{
    \setlength\tabcolsep{3pt} 
    \begin{tabular}{l | c c | c c}
     \hline
     Methods & \multicolumn{2}{c|}{Seq. 09} & \multicolumn{2}{c}{Seq. 10} \\
     & $t_{err}$ ($\%$) & $r_{err}$ ($^{\circ}/100m$) & $t_{err}$ ($\%$) & $r_{err}$ ($^{\circ}/100m$) \\
    \hline
    ORB-SLAM2 ~\cite{mur2017orb} & X & X & X & X \\
    \hline
    Zhou~\etal~\cite{zhou2017unsupervised} & 61.24 & 18.32 & 38.94 & 19.62 \\
    Depth-VO-Feat \cite{zhan2018unsupervised} & 42.33 & 11.88 & 25.83 & 11.58 \\
    CC \cite{ranjan2019competitive} & 51.45 & 14.39 & 34.97 & 17.09 \\
    SC-SfMLearner \cite{bian2019unsupervised} & 59.32 & 17.91 & 42.25 & 21.04 \\
    Ours & \textbf{7.72}  & \textbf{1.14} & \textbf{17.30} & \textbf{5.94} \\
    \hline
    \end{tabular}
    }
    \vspace{0.08cm}
    \caption{Visual odometry results on sampled sequence 09 and 10 with stride 4. The average translation and rotation errors are reported.}
    \vspace{-5pt}
\label{tab::vo-kitti-s4}
\end{table}
\begin{figure}[tb]
\scriptsize
\begin{tabular}{cc}
{\includegraphics[width=0.55\linewidth,height=120pt]{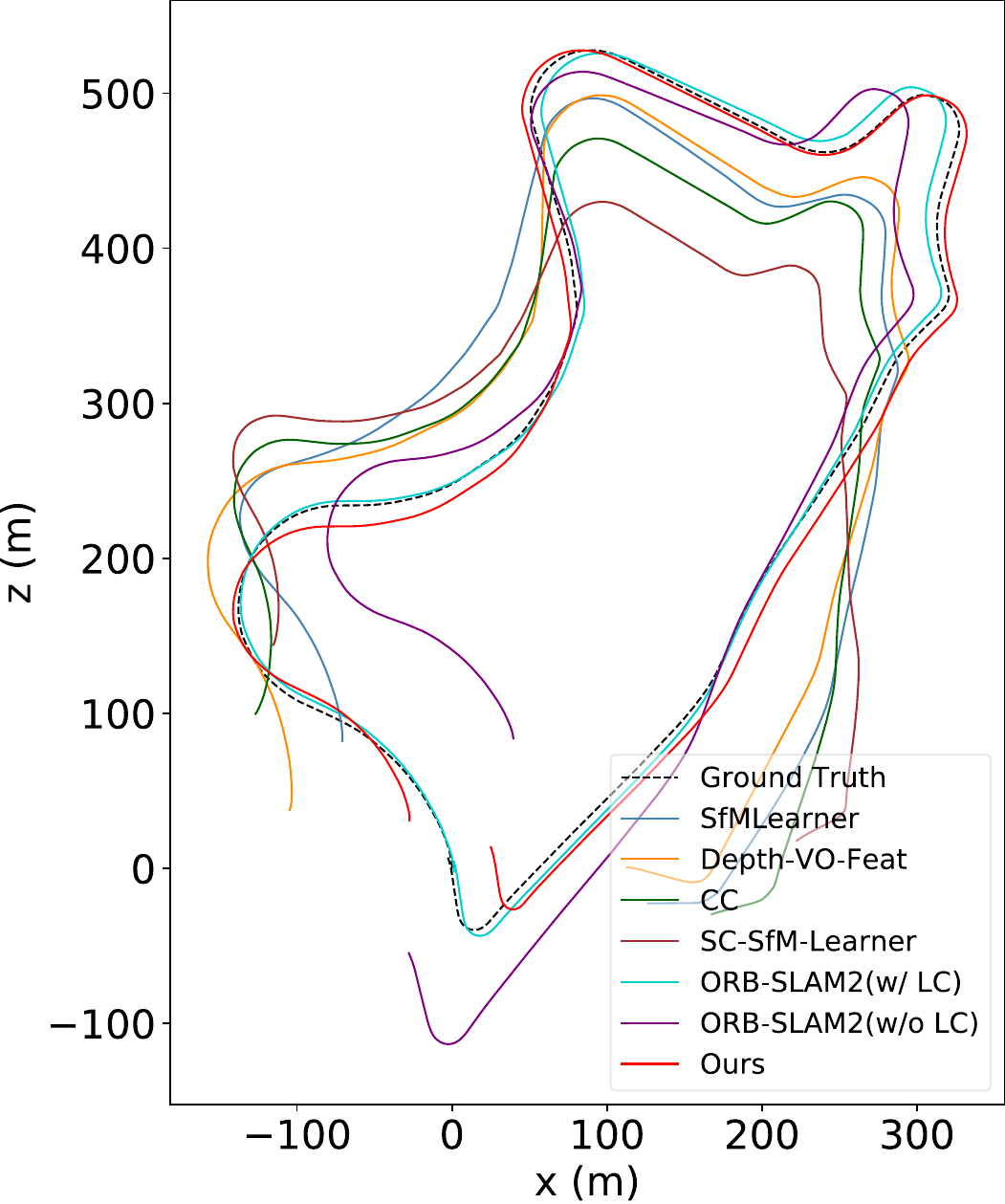}} &
{\includegraphics[width=0.35\linewidth, height=120pt]{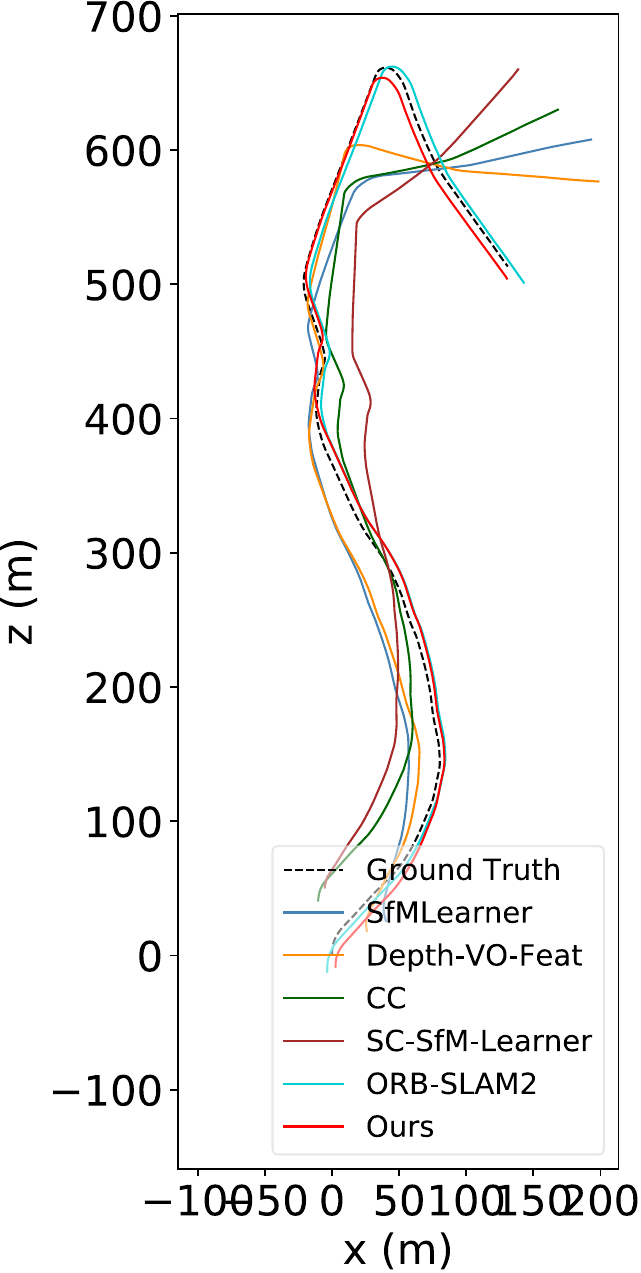}} \\

\end{tabular}
\centering
\caption{Visual odometry results on sampled sequence 09 and 10 with stride 2.}
\label{fig:supp_odo_s2}
\vspace{0pt}
\end{figure}

\begin{figure}[tb]
\scriptsize
\begin{tabular}{cc}
{\includegraphics[width=0.55\linewidth,height=120pt]{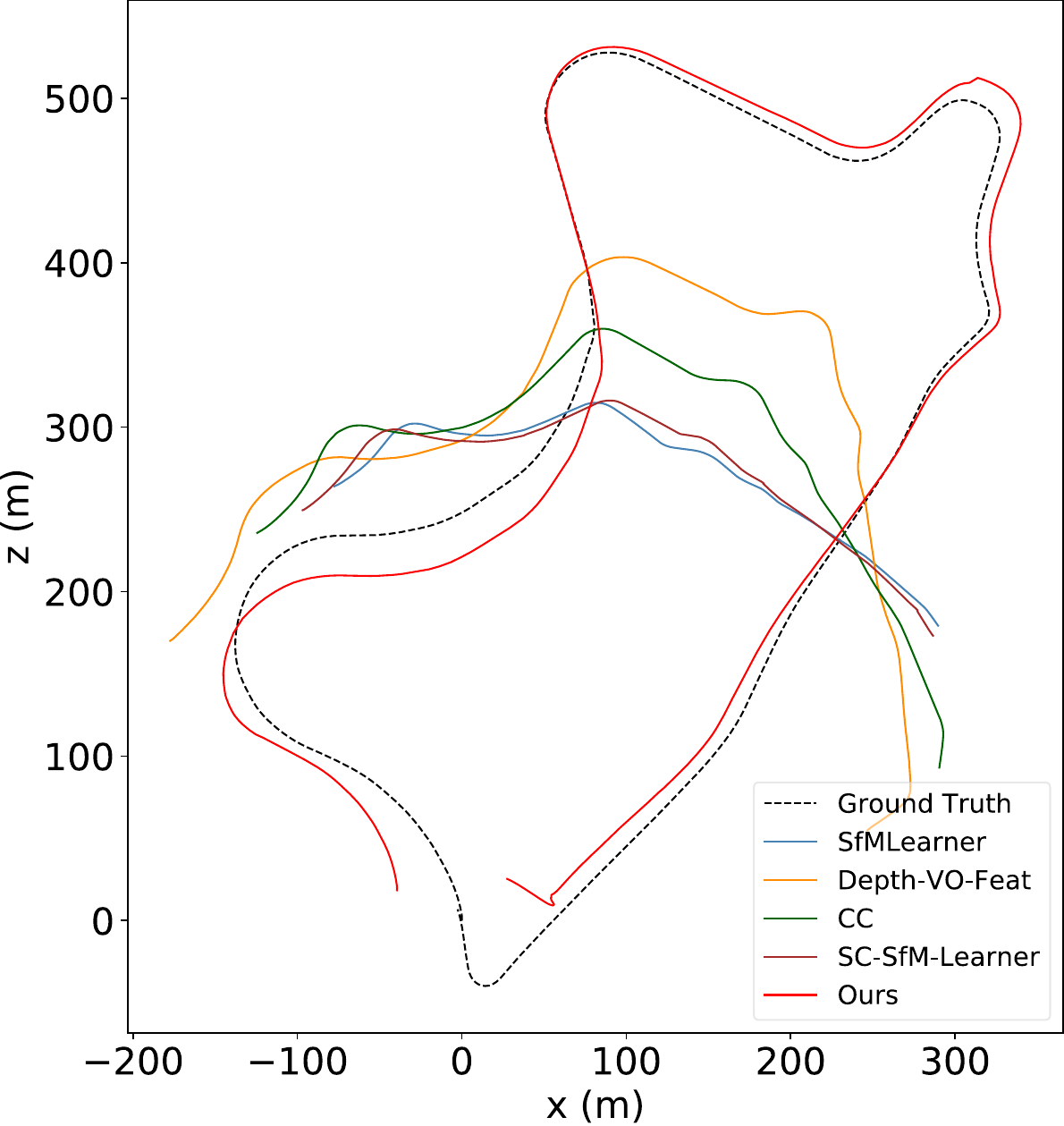}} &
{\includegraphics[width=0.35\linewidth, height=120pt]{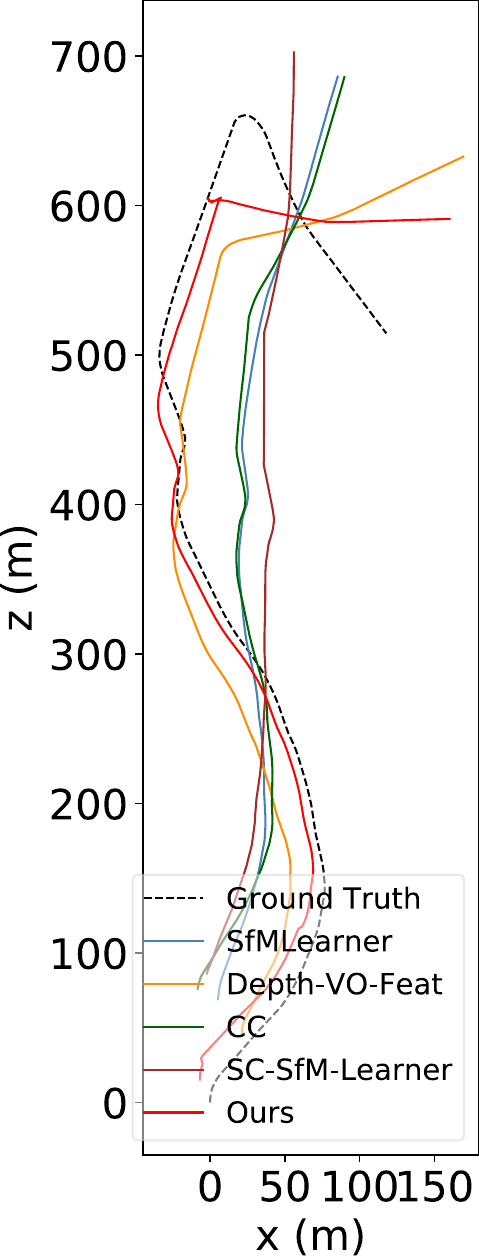}} \\

\end{tabular}
\centering
\caption{Visual odometry results on sampled sequence 09 and 10 with stride 4.}
\label{fig:supp_odo_s4}
\vspace{0pt}
\end{figure}

\section{Additional Comparison on sampled KITTI Odometry dataset}
\label{sec:odo}

To better demonstrate the robustness of our system, we provide additional comparison on sampled KITTI Odometry dataset. The test sequences 09 and 10 are sampled with stride 2 and 4, and we run the PoseNet-based learning systems and ORB-SLAM2 on these sampled sequences without additional training. Table \ref{tab::vo-kitti-s2} and \ref{tab::vo-kitti-s4} summarize the results of sampling with stride 2 and 4 respectively. Trajectories results are shown in Figure \ref{fig:supp_odo_s2} and \ref{fig:supp_odo_s4}. Again our system shows improved robustness and generalization ability compared to our baselines. However, when the camera moves extremely fast, such as sampling with stride 4 or more, the optical flow estimation becomes bottleneck and the performance degrades due to inaccurate correspondences.

\section{Numerical Results of TUM-RGBD dataset}
\label{sec:tum}
\begin{table}[tb]
\centering
    
    \scalebox{0.75}{
    \setlength\tabcolsep{3pt} 
    \begin{tabular}{l | c c c c}
     \hline
     Sequences & fr3/cabinet & fr2/desk & fr3/str\_ntex\_far & fr3/str\_tex\_far \\
    \hline
    PoseNet & 1.45 & 1.51 & 0.32 & 0.38 \\
    ORB-SLAM2 ~\cite{mur2017orb} & X & 0.006 & X & 0.009 \\
    Ours & 1.09 & 0.52 & 0.24 & 0.14 \\
    \hline
    \end{tabular}
    }
    \vspace{0.08cm}
    \caption{Results for selected sequences on TUM-RGBD dataset. We report the absolute translational RMSE in meter.}
    \vspace{-5pt}
\label{tab::supp-tum}
\end{table}

In Table \ref{tab::supp-tum}, we report the quantitative results of TUM-RGBD dataset. Our methods could produce reasonable trajectories under challenging scenarios while PoseNet baseline fails to generalize. ORB-SLAM2 relies on sparse ORB features to establish correspondences, and it suffers on large textureless regions (fr3/cabinet, fr3/str\_ntex\_far). However, ORB-SLAM2 works much better than ours when the scene contains rich textures (fr2/desk, fr3/str\_tex\_far). Our system could be further improved with better optical flow estimation and combination with back-end optimization. TUM-RGBD and NYUv2 are both indoor datasets and share some similar data distributions. We trained our method and PoseNet on TUM-RGBD dataset and directly tested on the NYUv2 dataset to demonstrate the transfer ability of trained model. Experimental results show that our model achieves better transfer performance (AbsRel 0.276) than PoseNet baseline (AbsRel 0.324). However, this transfer ability is still limited and has large room for improvement in the future.

\section{Additional Visualizations}
\label{sec:vis}
\begin{figure*}[tb]
\small
\setlength\tabcolsep{2.0pt} 
\renewcommand{\arraystretch}{4.0}
\begin{tabular}{ccc}
Image & Depth Estimation & Flow Estimation \\
{\includegraphics[width=0.33\linewidth]{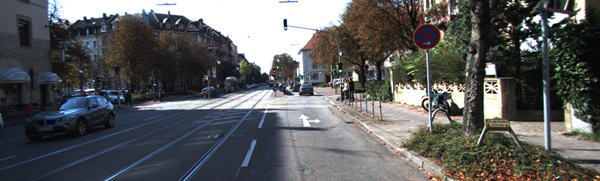}} &
{\includegraphics[width=0.33\linewidth]{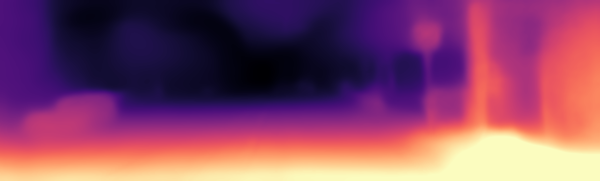}}
&
{\includegraphics[width=0.33\linewidth]{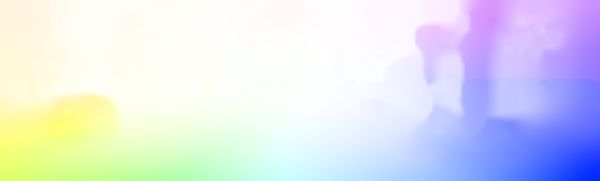}}
\\

{\includegraphics[width=0.33\linewidth]{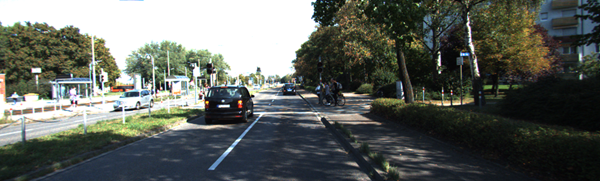}} &
{\includegraphics[width=0.33\linewidth]{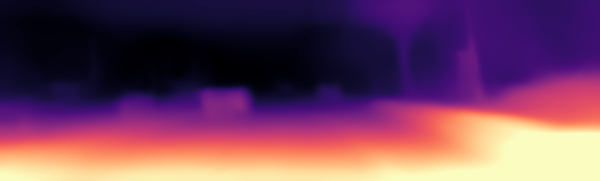}}
&
{\includegraphics[width=0.33\linewidth]{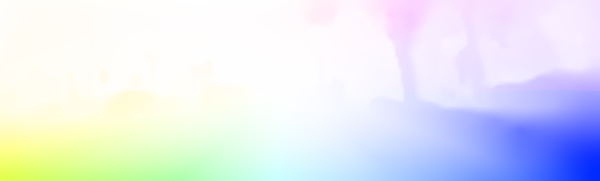}}
\\

{\includegraphics[width=0.33\linewidth]{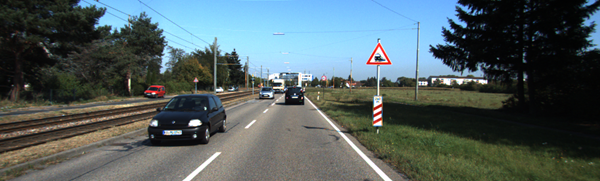}} &
{\includegraphics[width=0.33\linewidth]{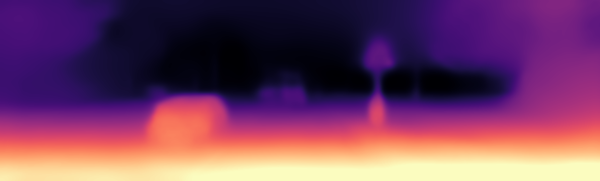}}
&
{\includegraphics[width=0.33\linewidth]{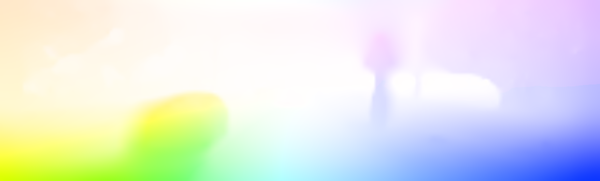}}
\\

{\includegraphics[width=0.33\linewidth]{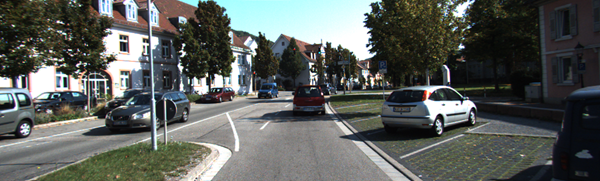}} &
{\includegraphics[width=0.33\linewidth]{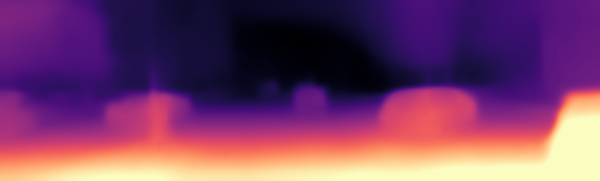}}
&
{\includegraphics[width=0.33\linewidth]{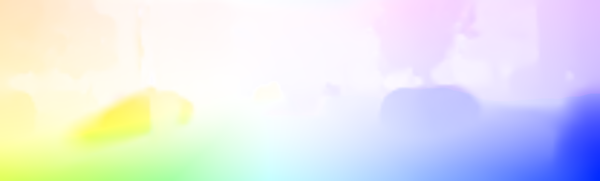}}
\\

{\includegraphics[width=0.33\linewidth]{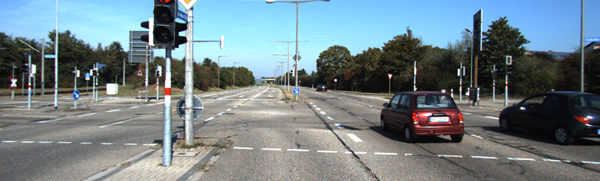}} &
{\includegraphics[width=0.33\linewidth]{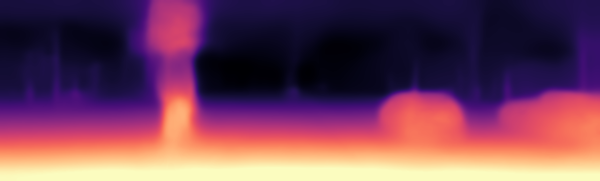}}
&
{\includegraphics[width=0.33\linewidth]{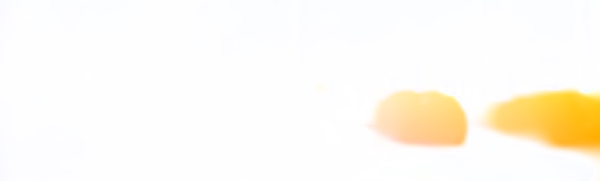}}
\\

{\includegraphics[width=0.33\linewidth]{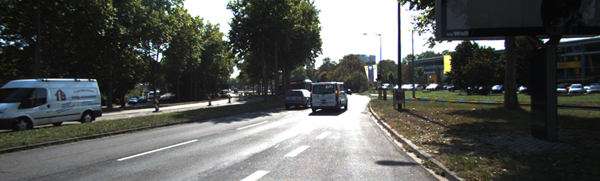}} &
{\includegraphics[width=0.33\linewidth]{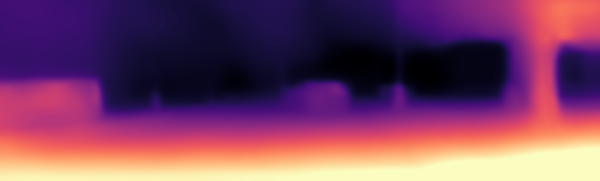}}
&
{\includegraphics[width=0.33\linewidth]{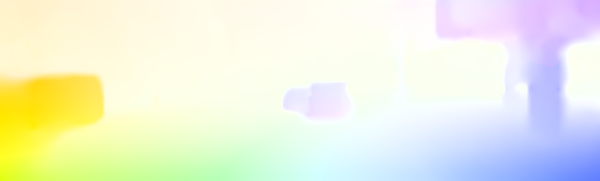}}
\\

{\includegraphics[width=0.33\linewidth]{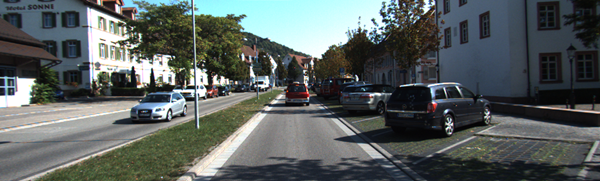}} &
{\includegraphics[width=0.33\linewidth]{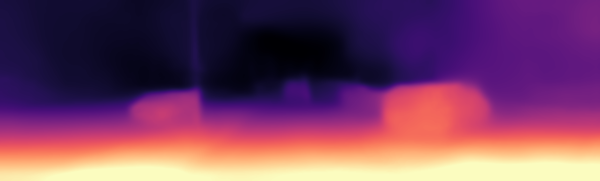}}
&
{\includegraphics[width=0.33\linewidth]{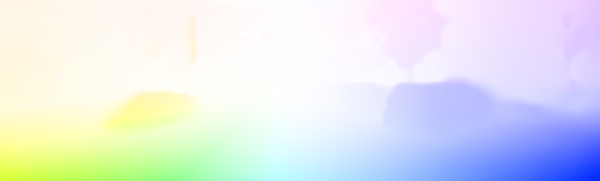}}
\\

{\includegraphics[width=0.33\linewidth]{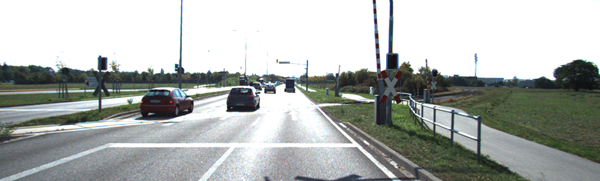}} &
{\includegraphics[width=0.33\linewidth]{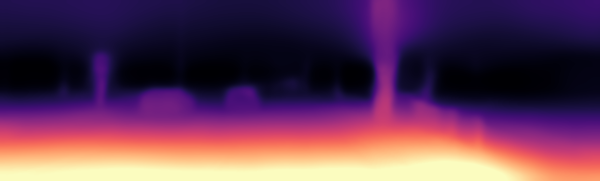}}
&
{\includegraphics[width=0.33\linewidth]{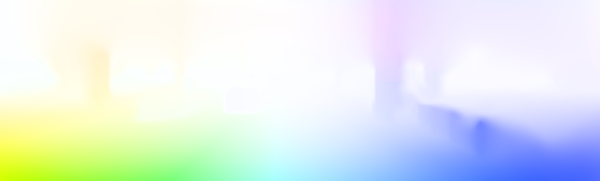}}
\\

\end{tabular}
\centering
\vspace{0.05cm}
\caption{Visualization for KITTI depth and flow estimation.
}
\label{fig::supp_kitti}
\vspace{0pt}
\end{figure*}

\begin{figure*}[tb]
\small
\setlength\tabcolsep{5pt} 
\renewcommand{\arraystretch}{2.0}
\begin{tabular}{cccc}
Image & Baseline & Ours & Groundtruth \\
{\includegraphics[width=0.20\linewidth]{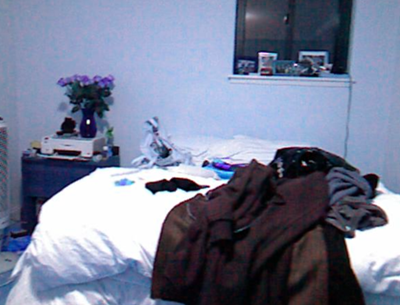}}
&
{\includegraphics[width=0.20\linewidth]{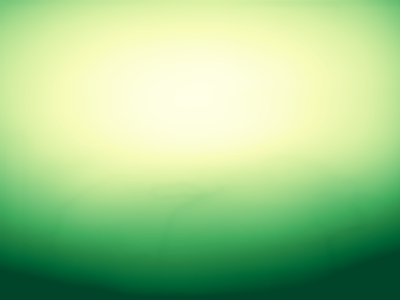}}
&
{\includegraphics[width=0.20\linewidth]{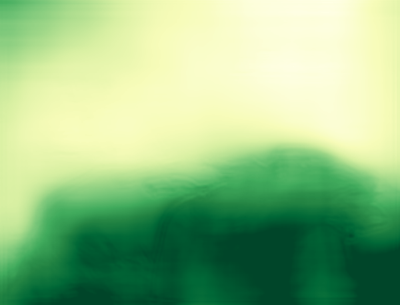}}
&
{\includegraphics[width=0.20\linewidth]{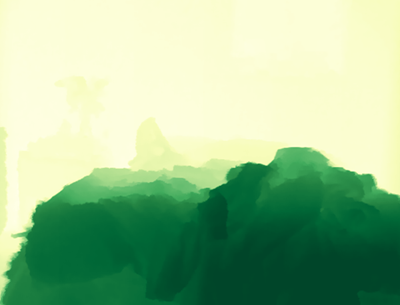}}
\\
{\includegraphics[width=0.20\linewidth]{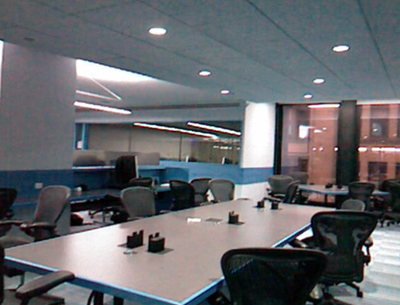}} &
{\includegraphics[width=0.20\linewidth]{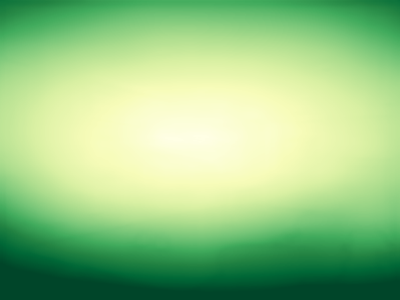}}
&
{\includegraphics[width=0.20\linewidth]{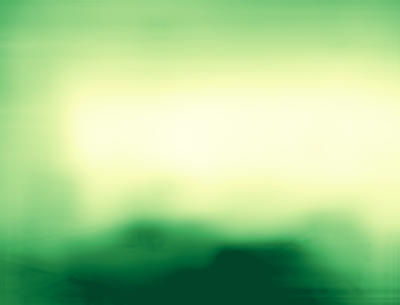}}
&
{\includegraphics[width=0.20\linewidth]{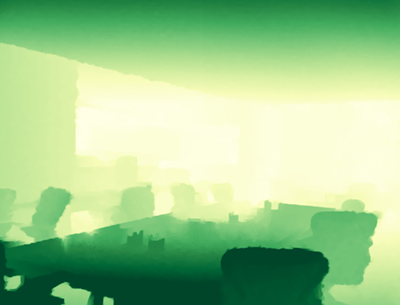}}

\\
{\includegraphics[width=0.20\linewidth]{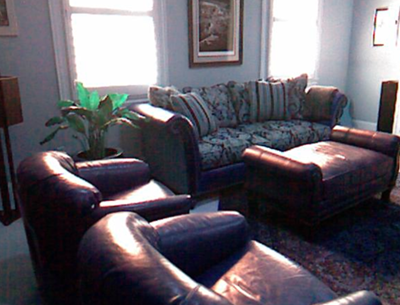}}
&
{\includegraphics[width=0.20\linewidth]{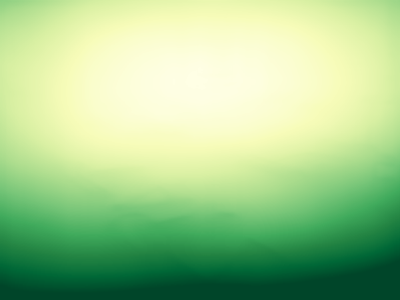}}
&
{\includegraphics[width=0.20\linewidth]{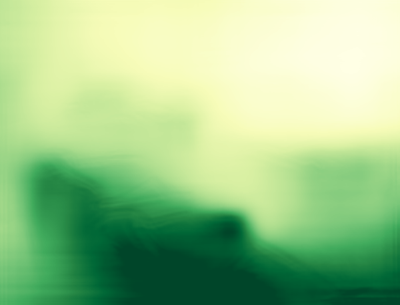}}
&
{\includegraphics[width=0.20\linewidth]{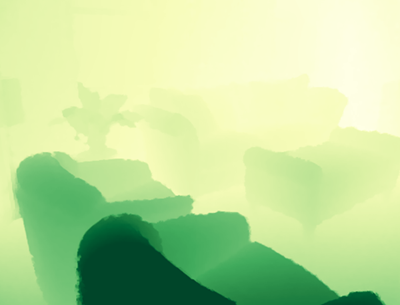}}
\\
{\includegraphics[width=0.20\linewidth]{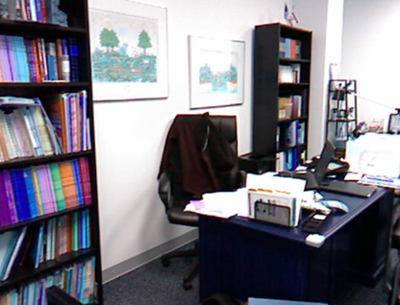}} &
{\includegraphics[width=0.20\linewidth]{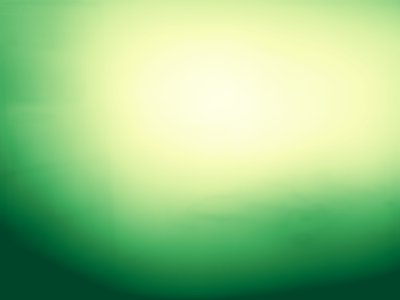}}
&
{\includegraphics[width=0.20\linewidth]{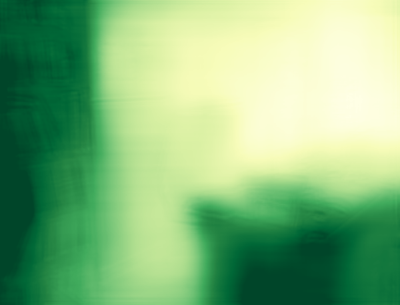}}
&
{\includegraphics[width=0.20\linewidth]{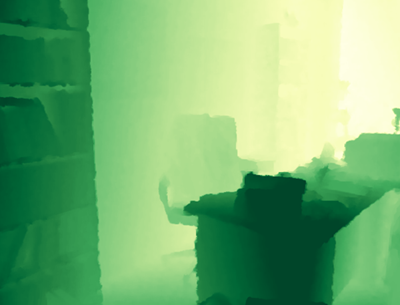}}

\\
{\includegraphics[width=0.20\linewidth]{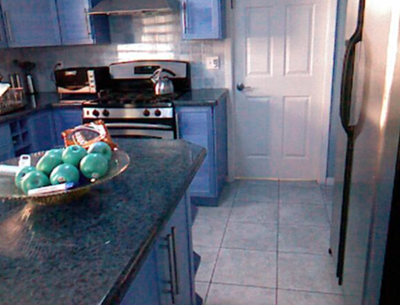}}
&
{\includegraphics[width=0.20\linewidth]{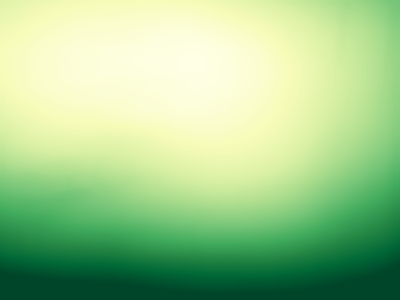}}
&
{\includegraphics[width=0.20\linewidth]{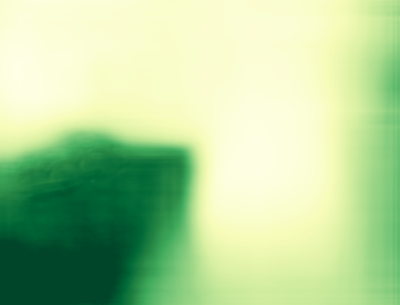}}
&
{\includegraphics[width=0.20\linewidth]{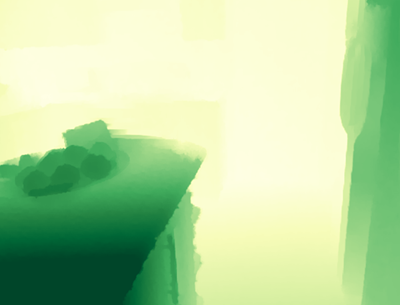}}
\\
{\includegraphics[width=0.20\linewidth]{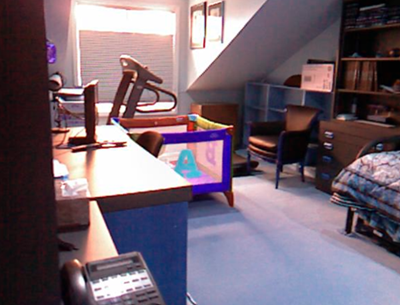}} &
{\includegraphics[width=0.20\linewidth]{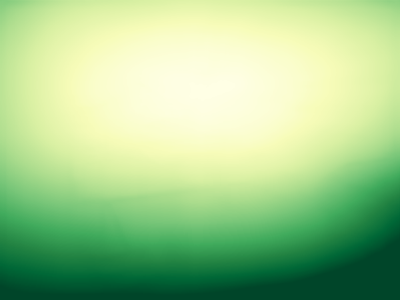}}
&
{\includegraphics[width=0.20\linewidth]{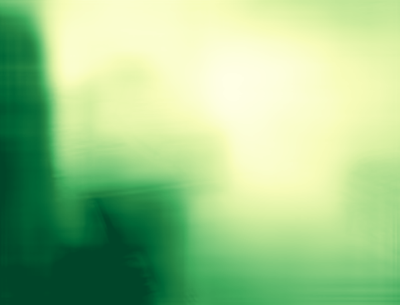}}
&
{\includegraphics[width=0.20\linewidth]{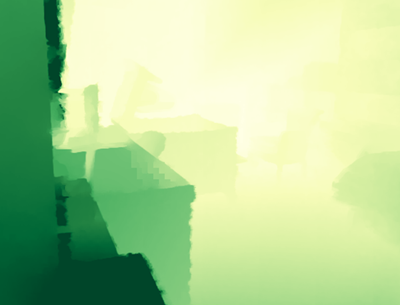}}

\\
{\includegraphics[width=0.20\linewidth]{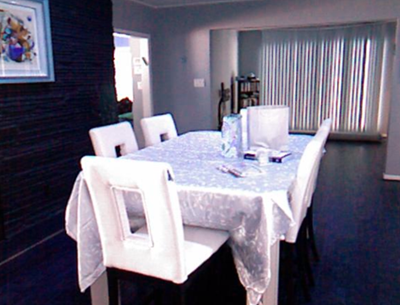}}
&
{\includegraphics[width=0.20\linewidth]{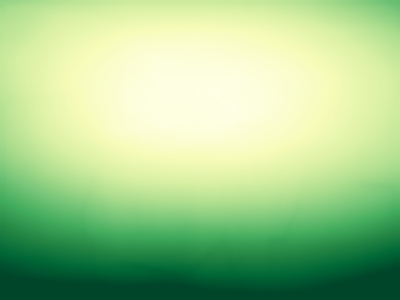}}
&
{\includegraphics[width=0.20\linewidth]{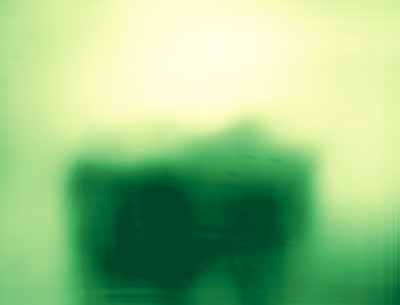}}
&
{\includegraphics[width=0.20\linewidth]{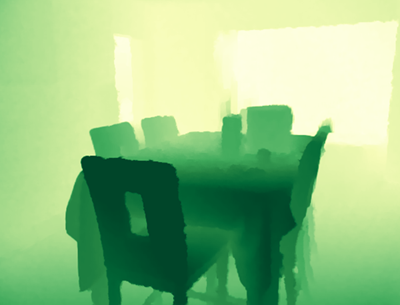}}
\\

\end{tabular}
\centering
\vspace{0.05cm}
\caption{Visualization for NYUv2 depth estimation. Baseline indicates replacing flow and triangulation module with PoseNet in our system. }
\label{fig::supp_nyu}
\vspace{0pt}
\end{figure*}

We provide more qualitative results on KITTI and NYUv2 dataset in Figure \ref{fig::supp_kitti} and Figure \ref{fig::supp_nyu}.

~\\

\end{document}